\documentclass{article}

\usepackage{style/unites}
\usepackage[utf8]{inputenc}
\usepackage[T1]{fontenc}
\usepackage{XCharter}
\usepackage[scaled=1.1]{zlmtt}
\usepackage{hyperref}
\usepackage{url}
\usepackage{booktabs}
\usepackage{graphicx}
\usepackage{amsfonts}
\usepackage{nicefrac}
\usepackage{amsmath,amssymb,amsthm}
\usepackage{marvosym}
\usepackage[most]{tcolorbox}
\usepackage{enumitem}
\usepackage{pifont}
\usepackage{adjustbox}
\usepackage{multirow}
\usepackage{tabularx}
\usepackage[compatibility=false]{caption}
\usepackage{subcaption}
\usepackage{setspace}
\usepackage{titlesec}
\usepackage{titletoc}
\usepackage[bottom]{footmisc}
\usepackage{microtype}
\definecolor{CarolinaUltraLight}{HTML}{E8F4F8}
\definecolor{citeblue}{RGB}{0,113,188}
\definecolor{firebrick}{rgb}{0.698,0.133,0.133}

\titlespacing\section{0pt}{4pt plus 4pt minus 2pt}{-2pt plus 2pt minus 2pt}
\titlespacing\subsection{0pt}{2pt plus 4pt minus 2pt}{-2pt plus 2pt minus 2pt}
\titlespacing\subsubsection{0pt}{2pt plus 4pt minus 2pt}{-2pt plus 2pt minus 2pt}

\newtcolorbox{titleblock}{
  enhanced,
  frame hidden,
  colback=CarolinaUltraLight,
  colframe=CarolinaUltraLight,
  boxrule=0pt,
  arc=10pt,
  left=14pt, right=14pt, top=14pt, bottom=14pt,
  width=\linewidth,
  before skip=12pt plus 4pt,
  after skip=12pt plus 4pt,
  grow to left by=1.5pt,
  grow to right by=1.5pt,
  before upper={\setlength{\parindent}{0pt}\setlength{\parskip}{0.5cm}}
}
\renewenvironment{abstract}{\normalfont\ignorespaces}{\par}

\newcommand{\bu}[1]{\textbf{\underline{#1}}}

\newcommand{\msUTSpr}{0.57}      % MSCAD UTS VUS-PR
\newcommand{\msMTSpr}{0.47}      % MSCAD MTS VUS-PR
\newcommand{\dUTSpr}{0.05}       % Δ vs CrossAD-tuned on UTS
\newcommand{\dMTSpr}{0.04}       % Δ vs CrossAD-no-tuning on MTS
\newcommand{\relUTSpr}{9.6\%}     % relative gain on UTS
\newcommand{\relMTSpr}{9.3\%}     % relative gain on MTS
\newcommand{\nBS}{50}            % number of baselines compared

\newcommand{\papertitle}{No Scale Left Behind: Multi-Scale Autoencoders with Bidirectional Attention for Time Series Anomaly Detection}
\newcommand{\paperrunningtitle}{No Scale Left Behind: Multi-Scale Autoencoders with Bidirectional Attention for TSAD}
\hypersetup{
  pdftitle={\papertitle},
  pdfauthor={Jiaheng Guo, Haochen Zhang, Morris Yu-Chao Huang, Jinhao Duan, Nicholas Konz, Tianlong Chen},
  pdfsubject={Time series anomaly detection},
  pdfpagemode=UseNone,
  colorlinks=true,
  linkcolor=firebrick,
  urlcolor=firebrick,
  citecolor=citeblue,
}
\icmlpreprint{Preprint.}

\begin{document}

% Reserve the height required by the template's first-page logo.
\addtolength{\topmargin}{-17pt}
\setlength{\headheight}{27pt}
\pagestyle{fancy}
\fancyhf{}
\renewcommand{\headrulewidth}{1pt}
\chead{\small\bfseries\paperrunningtitle}
\cfoot{\thepage}
\fancypagestyle{fancytitlepage}{%
  \fancyhf{}
  \lhead{\includegraphics[height=0.8cm]{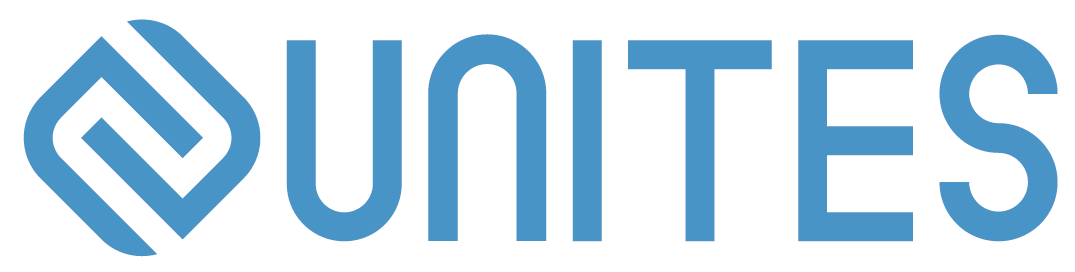}}
  \rhead{\itshape\today}
}
\thispagestyle{fancytitlepage}

\vspace*{0.5em}
\noindent
\begin{titleblock}
  {\setlength{\parskip}{0pt}\raggedright
   \setstretch{1.2}\LARGE\bfseries\papertitle\par}
  \vskip 0.2cm
  {\setlength{\parskip}{0pt}
   \begin{icmlauthorlist}
     \mbox{Jiaheng Guo}, \mbox{Haochen Zhang}, \mbox{Yu-Chao Huang},\
     \mbox{Jinhao Duan}, \mbox{Nicholas Konz}, \mbox{Tianlong Chen}
   \end{icmlauthorlist}\par
  \vskip 0.2cm
  UNITES Lab, University of North Carolina at Chapel Hill\par
  }
  \vskip 0.1cm
  % Original draft, written by Jiaheng
% \begin{abstract}
% Time Series Anomaly Detection plays a crucial role in various fields including healthcare, sensor monitoring, and finance. We observe that the anomalies can be categorized into different types, while most existing methods can only handle one of these types. We propose \bu{M}ulti-\bu{S}cale autoencoder with parallel \bu{C}ross-scale \bu{A}ttention for time series anomaly \bu{D}etection (\textbf{MSCAD}), a semi-supervised framework that leverages patches of different sizes to overcome the challenge, and a symmetric cross-attention mechanism to enable communication across different scales. On the comprehensive TSB-AD benchmark, we achieve state-of-the-art performance across all metrics against statistical baselines and modern methods by a large margin, with a VUS-PR score of 0.5473 on univariate time-series and a VUS-PR score of 0.4549 on multivariate time-series.
% \end{abstract}

% Original draft modified by Claude, then modified again by Jiaheng
\begin{abstract}
Time series anomaly detection (TSAD) plays a crucial role in healthcare, finance, industrial monitoring, and other sectors. Within and between these settings, anomalies span vastly different temporal scales, from sub-second point spikes to multi-hour drift patterns. 
However, most existing TSAD methods commit to a single temporal granularity, and multi-scale designs either analyze different scales in isolation or are constrained to a predefined coarse-to-fine hierarchy, both failing to sufficiently capture multi-scale interactions.
% that can obscure fine-scale anomalies without a coarse-scale correlate. 
%
To resolve this limitation, we propose \bu{M}ulti-\bu{S}cale Autoencoder with \bu{C}ross-Scale \bu{A}ttention for TSA\bu{D} (\textbf{MSCAD}), a simple yet powerful semi-supervised TSAD framework founded on parallel autoencoder branches corresponding to different patch sizes. 
% We propose \bu{M}ulti-\bu{S}cale Autoencoder with \bu{C}ross-Scale \bu{A}ttention for time series anomaly \bu{D}etection (\textbf{MSCAD}) to enable bidirectional information interaction across different scales. 
%
A stack of symmetric bidirectional cross-scale attention blocks enables every pair of scales to exchange information before reconstruction \textit{without} allowing any single scale to be privileged. 
% Anomalies within time series significantly vary in temporal extent, motivating anomaly detectors that combine information across multiple granularities. However, processing scales independently leaves their representations disconnected, and coarse-to-fine hierarchies restrict how fine-scale evidence can inform coarser representations. We introduce \textbf{MSCAD}, a multi-scale autoencoder for semi-supervised time series anomaly detection that enables bidirectional exchange between scales before reconstruction. Trained on anomaly-free data, MSCAD processes each fixed-length input window through parallel patch transformer branches with different patch sizes. A symmetric cross-scale attention bridge allows each branch to query representations from all other scales, and uniform averaging combines their reconstruction errors into anomaly scores.
%
On the comprehensive TSB-AD benchmark (40 datasets, 530 series), MSCAD achieves large performance gains against $\nBS$ baselines across multiple metrics, with VUS-PR of $\msUTSpr$($+\relUTSpr$) on the univariate split and $\msMTSpr$($+\relMTSpr$) on the multivariate split compared to the state-of-the-art.
\end{abstract}

\end{titleblock}
\blfootnote{%
\textsuperscript{\Letter} Corresponding authors: \texttt{jiaheng@unc.edu}, \texttt{\{nickk124,tianlong\}@cs.unc.edu}.
\\[2.5em]
\textit{\csname @icmlpreprint\endcsname}%
}

\section{Introduction}\label{sec:intro}
 Time series data underpins many critical sectors, such as medical diagnostics, industrial infrastructure, and finance, where the detection of anomalies is paramount. These irregularities vary significantly in nature, including physiological spikes in EEG/ECG signals, sudden industrial service outages, or gradual financial drifts. Effectively identifying these behaviors via time series anomaly detection (TSAD) is a fundamental requirement for ensuring the safe and reliable operation of these systems.

\begin{figure}[t]
  \centering
  
  \begin{subfigure}{0.9\linewidth}
    \centering
    \includegraphics[width=\linewidth]{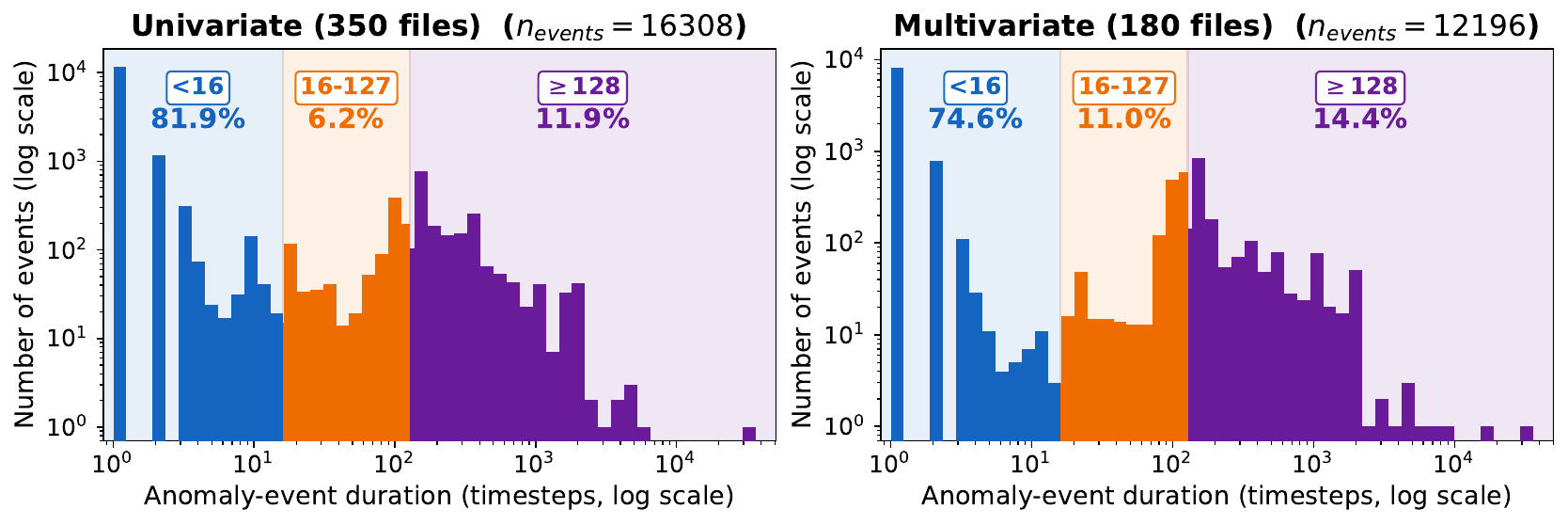}
  \end{subfigure}

  \begin{subfigure}{0.9\linewidth}
    \centering
    \includegraphics[width=\linewidth]{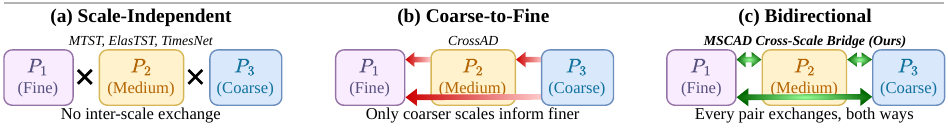}
  \end{subfigure}

  \caption{\textit{(Top)} \textbf{Anomaly-event duration distribution on TSB-AD.} Short events ($<16$ timesteps) dominate; long events ($\geq 128$ timesteps) form a non-trivial $12$--$14\%$ tail, motivating a multi-scale design. \textit{(Bottom)} \textbf{Cross-scale coupling topologies.} Existing multi-scale TSAD methods either keep scales independent or impose asymmetric coarse-to-fine flow. MSCAD uses a bidirectional bridge, allowing each scale to exchange information with every other scale symmetrically.}
  \label{fig:intro}
\end{figure}

Semi-supervised TSAD is the primary paradigm for high-stakes applications because labeled anomalies are notoriously scarce and diverse, necessitating models that learn the manifold of normalcy from anomaly-free training data. Over the past decade, research has produced various architectural backbones, including reconstruction-based detectors, forecasting-based approaches, and representation-based methods \citep{audibert2020usad, wu2023timesnet, yang2023dcdetector}. However, most of these frameworks treat temporal scale as a fixed hyperparameter, which limits their ability to capture both point-wise and trend-level irregularities simultaneously. Recent meta-analyses on the TSB-AD benchmark underscore this bottleneck, reporting that the performance gap between disparate backbones is surprisingly narrow \citep{liu2024elephant}. This suggests the field is restricted not by backbone choice, but by a failure to effectively integrate multi-scale semantics, which can range from transient, sub-second spikes (e.g., sensor malfunctions) to persistent, multi-hour regime shifts (Figure~\ref{fig:intro}).

A detector restricted to a single temporal receptive field faces an inherent trade-off, where a resolution that is too fine may overlook slow structural changes and one that is too coarse will smooth over transient events. While multi-scale designs have attempted to mitigate this, existing coupling strategies generally fall into two suboptimal categories. First, scale-independent parallel branches process scales in isolation and fuse outputs only via score-level weighting or feature concatenation, which fails to capture anomalies requiring joint reasoning across scales, such as a fine-scale deviation that is only salient in the context of a shifted coarse-scale background \citep{wu2023timesnet, zhang2024mtst}. Second, asymmetric hierarchy approaches restrict cross-scale information flow to one direction, using coarse-scale context as a conditioning signal for finer scales \citep{li2025crossad, chen2024pathformer}. CrossAD \citep{li2025crossad} makes this design explicit, utilizing block-diagonal masks that forbid cross-scale interaction inside the encoder, imposing a directed coarse-to-fine inductive bias that may be suboptimal when anomaly evidence is scale-local or reciprocal. We demonstrate that the strategy for coupling scales is as critical as the presence of multiple scales themselves, representing a fundamental gap in current architectural designs.
% To our knowledge, no prior TSAD method supports \emph{symmetric bidirectional} information exchange between parallel per-scale branches with no privileged reference.

% \begin{figure}
%     \centering
%     \includegraphics[width=\linewidth]{figs/intro.png}
%     \caption{Anomaly-event durations across the TSB-AD univariate evaluation split. The distribution is bimodal: $\sim75\%$ of events are shorter than $4$ timesteps (point anomalies), $\sim 12$--$15\%$ exceed the $128$-timestep window (long contextual anomalies), and the middle range carries $<4\%$ of events. Vertical dashed lines mark our patch sizes $P\in\{4,16,64\}$; the dotted line marks the window $W{=}128$.} % The two modes motivate using both the smallest scale ($P{=}4$, point coverage) and the largest scale ($P{=}64$, long-context coverage) simultaneously, with $P{=}16$ filling the sparse middle.
%     \label{fig:intro}
%     \vspace{-1em}
% \end{figure}

We close this gap with \textbf{MSCAD} (\bu{M}ulti-\bu{S}cale Autoencoder with \bu{C}ross-Scale \bu{A}ttention for Time Series Anomaly \bu{D}etection). Each input window is processed in parallel by three autoencoder branches for different patch sizes. Crucially, we use a small stack of \textit{symmetric cross-scale attention bridges} between the per-scale encoders and decoders so that every scale's encoded tokens cross-attend to the concatenation of all other scales' tokens, with no hierarchy or privileged reference scale, so each pair of scales freely exchanges information in both directions before reconstruction. These bridged representations are then decoded at each scale, and the resulting per-scale reconstruction errors are fused via a simple uniform aggregator. Surprisingly, we find that this straightforward aggregation approach matches or outperforms every learned input-adaptive gating mechanism we evaluated, indicating that the preceding cross-scale attention mechanism already effectively reconciles temporal semantics prior to reconstruction. On the comprehensive TSB-AD benchmark, MSCAD attains state-of-the-art VUS-PR of $\mathbf{\msUTSpr}$ on univariate and $\mathbf{\msMTSpr}$ on multivariate splits, with 3-seed standard deviation below $.002$ and no per-file hyperparameter tuning, representing a significant advance toward robust, general-purpose TSAD that maintains strong performance across diverse domains without the need for manual, dataset-specific configuration. % \jiaheng{Value TBU after multi-seed run. this current value is outdated: i have another config with higher performance now (dmodel=256, cross-scale depth=2, sum instead of attn, will update later if multiseed run proves stablity. also may need large modification regarding related contents.} 
Our main contributions are summarized as follows:
% \begin{itemize}[leftmargin=1.4em,topsep=2pt,itemsep=1pt]
%   \item \textbf{Problem identification.} Existing multi-scale TSAD either runs scales in isolation or imposes a coarse-to-fine hierarchy, both of which miss fine-scale anomalies without a coarse correlate (\S\ref{sec:related}). % if figure 1 can be out
%   \item \textbf{Symmetric cross-scale bridge.} A bidirectional mechanism enabling information exchange between all scale pairs without a privileged reference. We find that a parameter-free mean-pool aggregator matches or exceeds multi-head attention, demonstrating that the performance gain is driven by the interaction itself rather than operator complexity (\S\ref{sec:method}).
%   \item \textbf{Simplified fusion.} A uniform aggregator matches or beats every learned input-adaptive gate we tested, indicating routing is redundant once scales communicate symmetrically (\S\ref{sec:exp}).
%   \item \textbf{State-of-the-art on TSB-AD.} MSCAD achieves VUS-PR of $\msUTSpr$ (UTS) and $\msMTSpr$ (MTS) across 45 baselines.
% \end{itemize}
\begin{itemize}[leftmargin=1.4em,topsep=2pt,itemsep=1pt]
    \item We identify a structural limitation of existing multi-scale TSAD methods: they either process scales independently or impose a coarse-to-fine hierarchy, missing the symmetric, bidirectional exchange between every pair of scales.
    \item We propose MSCAD, a multi-scale autoencoder with a symmetric cross-scale bridge that performs bidirectional cross-attention between every pair of scales without a privileged reference scale.
    \item We show through ablations that symmetric exchange is the key design component: replacing it with a coarse-to-fine variant reduces the gain to within seed noise of a no-bridge baseline. A uniform aggregator matches or outperforms learned input-adaptive gates.
    \item MSCAD achieves state-of-the-art performance across multiple metrics on the TSB-AD benchmark, with VUS-PR of $\msUTSpr$ on TSB-AD-U and $\msMTSpr$ on TSB-AD-M across $\nBS$ baselines.
\end{itemize}

\section{Related Work}\label{sec:related}

\paragraph{Semi-supervised time series anomaly detection.}
TSAD under the semi-supervised setting has been approached from several broad angles. \emph{Reconstruction-based} methods train a model on anomaly-free data and score anomalies by reconstruction error: USAD~\citep{audibert2020usad} introduces adversarial training of paired autoencoders; Anomaly Transformer~\citep{xu2022anomalytransformer} exploits the asymmetric \emph{prior-association / series-association} gap within a single-scale Transformer; DCdetector~\citep{yang2023dcdetector} contrasts dual patch-attention views of the same series. \emph{Forecasting- and frequency-based} methods instead score deviations from predicted or frequency-domain normal patterns: TimesNet~\citep{wu2023timesnet} converts 1-D series into 2-D tensors along dominant periods; FITS~\citep{xu2024fits} models time series with a lightweight frequency interpolation module; CATCH~\citep{wu2025catch} introduces channel-aware frequency patching for multivariate anomaly detection; KAN-AD~\citep{zhou2025kanad} uses Fourier-based Kolmogorov--Arnold networks to model smooth temporal patterns with few parameters. Recent \emph{general and representation-based} detectors further improve robustness and transferability: DADA~\citep{shentu2025dada} pre-trains a general anomaly detector with adaptive bottlenecks and dual adversarial decoders, while PaAno~\citep{paano2026} learns patch-level representations with a lightweight 1D-CNN and scores anomalies by comparing test patches against normal training patches. % Across these design axes, the above methods do not provide symmetric representation-level exchange among parallel temporal scales, which is the gap MSCAD targets.

\paragraph{Multi-scale anomaly detectors.}
The natural response to the granularity dilemma is to run the detector at multiple scales. Existing multi-scale designs for TSAD fall into two camps, both structurally different from ours. \emph{(i) Scale-independent parallel branches.} MTST \citep{zhang2024mtst} and ElasTST \citep{liang2024elastst} tokenize at multiple patch sizes but process each scale independently and concatenate features without cross-scale attention; TimesNet \citep{wu2023timesnet} applies period-specific 2-D convolution and recombines only via amplitude-weighted summation; None of these allow one scale's representation to be shaped by another's within the backbone. \emph{(ii) Asymmetric hierarchies.} Besides CrossAD \citep{li2025crossad},
% explicitly enforces this asymmetry: its encoder use a block-diagonal attention mask that forbids cross-scale interaction, and its cross-scale decoder lets each scale attend only to \emph{coarser} scales, privileging the coarse scale as the unique conditioning signal. 
Pathformer \citep{chen2024pathformer} routes forecasting information across scales via an adaptive pathway gate; Scaleformer and coarse-to-fine pyramidal Transformers \citep{shabani2023scaleformer,fan2024perimidformer} refine predictions hierarchically; TimeMixer \citep{wang2024timemixer} mixes past and future decompositions with directed (non-symmetric) flows. All share the same privileged-scale inductive bias.
% \emph{departs from both camps}: we place a symmetric bidirectional cross-scale bridge between parallel per-scale encoders and per-scale decoders, so each scale's tokens can query and be queried by every other scale, with no privileged reference scale and no hierarchy. This 
% MSCAD is, to our knowledge, the first multi-scale TSAD architecture in which inter-scale information flow is genuinely symmetric at the representation level.
In contrast, MSCAD is designed to enable symmetric inter-scale information flow at the representation level, avoiding both scale-isolated processing and privileged coarse-to-fine hierarchies.

% \paragraph{Relevant benchmarks and metrics.}
% Early TSAD research standardized around a handful of real-world multivariate datasets (SMAP and MSL \citep{hundman2018smap}, SMD \citep{su2019omnianomaly}, SWaT and WADI \citep{goh2016swat}, PSM \citep{abdulaal2021psm}) and a growing suite of univariate collections (UCR Anomaly Archive \citep{wu2023ucr}, NAB, Yahoo). Evaluation on these datasets was long dominated by \emph{point-adjusted} F1, which Kim et al.~\citep{kim2022rigorous} showed can be spuriously inflated: random scores can exceed $0.70$ PA-F1 on most standard datasets.
% %, because the metric credits an entire anomaly segment as correctly detected if a single timestep within it crosses the threshold. 
% In response, the community has converged on \emph{volume-under-the-surface} metrics (VUS-PR and VUS-ROC \citep{paparrizos2022volume}) which integrate over threshold and temporal buffer jointly and are not gameable by score sharpness. The TSB-AD benchmark \citep{liu2024elephant} consolidates this shift: it enforces a tune-then-evaluate protocol, with it's evaluation set curating $350$ univariate and $180$ multivariate evaluation series spanning $40$ real-world datasets, and reports all TSAD methods on a unified VUS-PR--primary leaderboard with $45$ baselines. We adopt TSB-AD as our main benchmark and VUS-PR as primary metric for exactly these reasons.

% COMPRESS: two-third of current
\section{Method}\label{sec:method}

\subsection{Problem Formulation}\label{sec:method:problem}

We consider TSAD under the standard semi-supervised setting \citep{liu2024elephant}. Let $X = (x_1, \ldots, x_T) \in \mathbb{R}^{T \times C}$ denote a time series of length $T$ with $C \geq 1$ channels. 
% (univariate when $C = 1$, multivariate otherwise). 
The training split $\mathcal{X}^{\text{tr}}$ is assumed \textit{anomaly-free}; no anomaly labels are observed during training. At test time, given an unseen series $X^{\text{te}}\in\mathcal{X}^{\text{te}}$ of length $T^{\text{te}}$, the goal is to produce a per-timestep anomaly score $\mathbf{s} \in \mathbb{R}^{T^{\text{te}}}$, where larger $s_t$ indicates a higher likelihood that timestep $t$ is anomalous. Performance is measured against a ground-truth label sequence $\mathbf{y} \in \{0,1\}^{T^{\text{te}}}$ via threshold-free volume-under-the-surface (VUS) metrics \citep{paparrizos2022volume} (primarily VUS-PR), which integrate $\mathbf{s}$ against $\mathbf{y}$ over all detection thresholds and a small temporal tolerance.

\textbf{Sliding-window framework.} We adopt a standard sliding-window framework where both training and inference operate on overlapping windows of length $W$ extracted with stride $\tau$, i.e., 
\begin{equation}\label{eq:window}
    \mathbf{x}^{(i)} \;=\; X[\,i\tau : i\tau + W] \;\in\; \mathbb{R}^{W \times C}, \quad \forall i \in \{0, 1, \ldots,\lfloor (T - W)/\tau\rfloor\}.
\end{equation}

% \begin{equation}
%   \mathbf{x}^{(i)} \;=\; X\bigl[\,i\tau : i\tau + W\,\bigr] \;\in\; \mathbb{R}^{W \times C},
%   \qquad i = 0, 1, \ldots, \bigl\lfloor (T - W)/\tau \bigr\rfloor.
%   \label{eq:window}
% \end{equation}

A scoring model $f_\theta: \mathbb{R}^{W \times C} \to \mathbb{R}^{+}$ assigns a window-level score $f_\theta(\mathbf{x}^{(i)})$; per-timestep scores $s_t$ are recovered by averaging over all windows that cover timestep $t$. Following \citet{liu2024elephant}, channels are processed independently and per-channel scores are averaged across channels, allowing a univariate model to handle multivariate inputs without parameter inflation.

\textbf{Reconstruction-based scoring.} A reconstruction-based detector parameterizes $f_\theta$ via an autoencoder $g_\theta: \mathbb{R}^{W \times C} \to \mathbb{R}^{W \times C}$ trained on $\mathcal{X}^{\text{tr}}$ to minimize the objective $\mathcal{L}(\theta) \;=\; \mathbb{E}_{\mathbf{x} \sim \mathcal{X}^{\text{tr}}} \bigl\| \mathbf{x} - g_\theta(\mathbf{x}) \bigr\|_2^2$. 
% \begin{equation}
%   \mathcal{L}(\theta) \;=\; \mathbb{E}_{\mathbf{x} \sim X^{\text{tr}}} \bigl\| \mathbf{x} - g_\theta(\mathbf{x}) \bigr\|_2^2,
%   \label{eq:recon-loss}
% \end{equation}
and scores each test window by its squared reconstruction error, $f_\theta(\mathbf{x}) = \|\mathbf{x} - g_\theta(\mathbf{x})\|_2^2$. The premise is that $g_\theta$, fit only on normal traces, faithfully reconstructs in-distribution windows but fails on subsequences whose patterns deviate from training, so reconstruction error signals deviation. % MSCAD adopts this paradigm and decomposes $g_\theta$ into parallel branches over multiple temporal scales.

\begin{figure}[t]
  \centering
  \includegraphics[width=\linewidth]{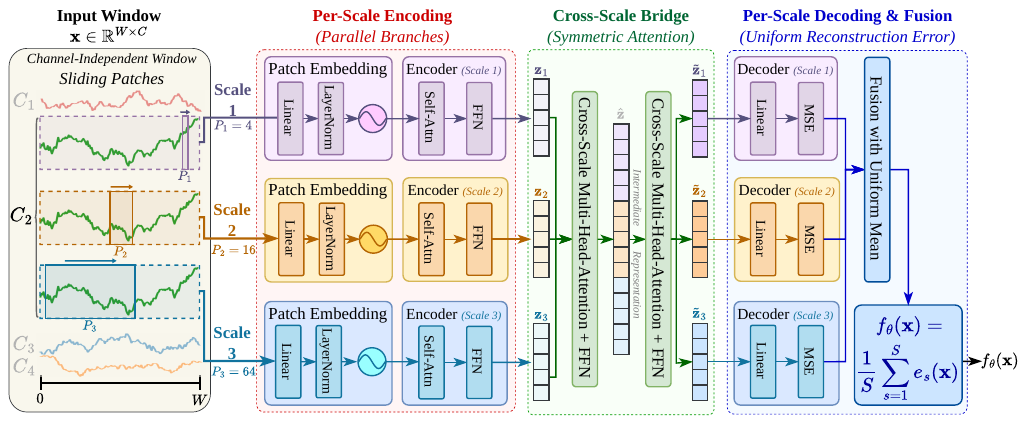}
  \caption{\textbf{MSCAD architecture.} Three parallel patch-Transformer branches at scales $\mathcal{P} = \{4, 16, 64\}$ (\S\ref{sec:method:branches}) are connected by a stack of $B$ symmetric cross-scale bridges (\S\ref{sec:method:bridge}); each branch decodes its bridged latents back to patch space, and the uniform mean of the per-scale reconstruction errors is the window score $f_\theta(\mathbf{x})$.}
  \label{fig:arch}
\end{figure}

\subsection{Framework Overview}\label{sec:method:overview}
The core of MSCAD is its reconstruction model $g_\theta$, a multi-scale autoencoder with symmetric bidirectional cross-scale communication. Given an input window $\mathbf{x} \in \mathbb{R}^{W \times C}$, the model produces the window-level anomaly score $f_\theta(\mathbf{x})$ end-to-end in the following three stages (Figure~\ref{fig:arch}).
\ding{172} \textbf{Per-scale encoding.} The window is processed in parallel by $S$ \emph{scale branches} indexed by a fixed set of patch sizes $\mathcal{P} = \{P_1, P_2, \ldots, P_S\}$ (we use $S=3$ and $\mathcal{P} = \{4, 16, 64\}$ throughout the paper). Each branch $\mathrm{Branch}_s$ tokenizes $\mathbf{x}$ into $N_s$ patch tokens via a strided patch embedding and processes them with a Transformer encoder, producing per-scale latents $\mathbf{z}_s \in \mathbb{R}^{N_s \times d}$ (\S\ref{sec:method:branches}).
\ding{173} \textbf{Symmetric cross-scale bridge.} The per-scale latents $\{\mathbf{z}_s\}_{s=1}^{S}$ pass through a stack of $B$ bridge modules that lets each scale's tokens attend to the concatenated tokens of all other scales, yielding context-mixed latents $\{\tilde{\mathbf{z}}_s\}_{s=1}^{S}$ of the same per-scale shapes. The bridge is \emph{symmetric} (\textit{i.e.}, no scale is privileged as a conditioning signal) and \emph{bidirectional} (\textit{i.e.}, every pair of scales both informs and is informed by the other), which is the core architectural departure from prior multi-scale TSAD (\S\ref{sec:method:bridge}).
\ding{174} \textbf{Per-scale decoding and uniform fusion.} Each bridged latent $\tilde{\mathbf{z}}_s$ is projected back to the patch space by a linear decoder, yielding a per-scale window-level reconstruction error $e_s(\mathbf{x}) \in \mathbb{R}_{\geq 0}$, the mean squared error between the reconstructed and original patches at scale $s$ (\S\ref{sec:method:branches}). The $S$ per-scale errors are fused into the final window-level anomaly score by a simple uniform averaging $f_\theta(\mathbf{x}) \;=\; \frac{1}{S} \sum_{s=1}^{S} e_s(\mathbf{x})$,
% \begin{equation}
%   f_\theta(\mathbf{x}) \;=\; \frac{1}{S} \sum_{s=1}^{S} e_s(\mathbf{x}),
%   \label{eq:fusion}
% \end{equation}
which also serves as the per-window training loss (\S\ref{sec:method:training}).

% \textbf{Design rationale.} 
Each of the three stages maps directly to the gap diagnosed in \S\ref{sec:intro}: parallel branches preserve scale-specific reconstruction targets so no scale absorbs another's signal, the symmetric bridge enables bidirectional inter-scale information flow without a coarse-to-fine asymmetry, and uniform fusion eschews learned routing, which we find redundant once the bridge has already reconciled semantics across scales (\S\ref{sec:exp:ablation}). % The remainder of \S\ref{sec:method} elaborates each component.

\subsection{Multi-Scale Patch Branches}\label{sec:method:branches}

Each scale branch $\mathrm{Branch}_s$ ($s = 1, \ldots, S$) is a lightweight autoencoder that operates at a single temporal granularity, defined by its patch size $P_s \in \mathcal{P}$. The encoder $\mathrm{Enc}_s$ produces the per-scale latent $\mathbf{z}_s$ to be utilized by the cross-scale bridge (\S\ref{sec:method:bridge}), and the decoder $\mathrm{Dec}_s$ reconstructs the original patches and yields the per-scale window-level error $e_s(\mathbf{x})$.

\textbf{Patch extraction and embedding.} Given a single-channel slice of an input window $\mathbf{x} \in \mathbb{R}^W$
% here for notational compactness, with multi-channel inputs handled channel-independently per \S\ref{sec:method:problem}—
the branch extracts overlapping patches with stride $P_s/2$ (50\% overlap): $$\mathbf{u}_s^{(j)} = \mathbf{x}\bigl[\,j \cdot P_s/2 : j \cdot P_s/2 + P_s
\,\bigr] \in \mathbb{R}^{P_s}, \quad j = 0, 1, \ldots, N_s - 1,$$ yielding $N_s = \lfloor 2(W - P_s)/P_s \rfloor + 1$ patches per window. Each patch is linearly projected to a $d$-dimensional token by a learnable patch embedding $\mathrm{PatchEmbed}_s : \mathbb{R}^{P_s} \to \mathbb{R}^d$, with a learnable absolute positional encoding $\mathbf{p}_s^{(j)} \in \mathbb{R}^d$ added on top: $\mathbf{h}_s^{(j)} = \mathrm{PatchEmbed}_s\bigl(\mathbf{u}_s^{(j)}\bigr) + \mathbf{p}_s^{(j)}.$

\textbf{Per-scale encoding.} The patch tokens $\mathbf{H}_s = \bigl(\mathbf{h}_s^{(0)}, \ldots, \mathbf{h}_s^{(N_s - 1)}\bigr) \in \mathbb{R}^{N_s \times d}$ are processed by a stack of $L$ standard Transformer encoder layers, using multi-head self-attention with $h$ heads and a GELU feed-forward sublayer of width $4d$, denoted as $\mathbf{z}_s = \mathrm{Enc}_s(\mathbf{H}_s) \in \mathbb{R}^{N_s \times d}.$ Crucially, the $S$ branches do not share parameters, ensuring that each scale can specialize its representation to satisfy its respective reconstruction objective without interference from other temporal granularities.
% without constraint from the others.

% \textbf{Choice of patch sizes.} Throughout the paper we use $\mathcal{P} = \{4, 16, 64\}$: three patch sizes spanning roughly an order of magnitude in temporal extent, chosen so that no patch at one scale completely covers a patch at the next. Smaller sets, such as $\{4, 64\}$, drop middle-scale coverage; larger sets, such as $\{4, 8, 16, 32, 64\}$, add parameters without consistent benefit. We validate this choice empirically in \S\ref{sec:exp:ablation} and report the full scale-count sweep in Appendix~\ref{app:ablations}. The reference configuration uses $L = 2$ Transformer layers per branch, $h = 4$ attention heads, and token dimension $d = 256$.

\textbf{Per-scale decoding.} The subsequent decoding operates on the bridged counterpart $\tilde{\mathbf{z}}_s$ (\S\ref{sec:method:bridge}) of $\mathbf{z}_s$. Each token $\mathbf{z}_s^{(j)}$ is mapped back to patch space by a single linear projection $\mathrm{Dec}_s : \mathbb{R}^d \to \mathbb{R}^{P_s}$:
$\hat{\mathbf{u}}_s^{(j)} = \mathrm{Dec}_s\bigl(\mathbf{z}_s^{(j)}\bigr), \quad j = 0, \ldots, N_s - 1.$
The per-scale window-level reconstruction error is the mean squared error between reconstructed and original patches, averaged over patch positions and within-patch timesteps:
$e_s(\mathbf{x}) = \frac{1}{N_s P_s} \sum_{j=0}^{N_s - 1} \| \hat{\mathbf{u}}_s^{(j)} - \mathbf{u}_s^{(j)} \|_2^2.$
This reconstruction error constitutes the per-scale signal for the aggregated window-level score and provides the local objective for the training process detailed in \S\ref{sec:method:training}.

\subsection{Symmetric Cross-Scale Bridge}\label{sec:method:bridge}

The cross-scale bridge is a unified module positioned between the per-scale encoders and decoders (\S\ref{sec:method:branches}). It accepts the $S$ per-scale latents $\{\mathbf{z}_s\}_{s=1}^S$ and produces transformed counterparts $\{\tilde{\mathbf{z}}_s\}_{s=1}^S$ of identical dimensions, ensuring that each $\tilde{\mathbf{z}}_s$ incorporates information from all other temporal scales. By default we utilize multi-head cross-attention with parameters shared across scales. Importantly, this design is both \textit{symmetric} (every scale simultaneously serves as both query and context) and \textit{bidirectional} (every pair of scales can exchange information within a single forward pass).

% \textbf{Cross-attention block.} For each scale $s$, the bridge forms a \emph{context sequence} by concatenating the latents of all other scales along the token axis,
% \begin{equation}
%   \mathbf{C}_s \;=\; \mathrm{concat}\bigl(\{\mathbf{z}_j : j \neq s\}\bigr) \;\in\; \mathbb{R}^{(\sum_{j \neq s} N_j) \times d},
%   \label{eq:context}
% \end{equation}
% and lets the query scale's tokens attend to it via standard multi-head attention with $h$ heads:
% \begin{equation}
%   \mathbf{a}_s \;=\; \mathrm{MHA}\bigl( Q = \mathbf{z}_s,\; K = \mathbf{C}_s,\; V = \mathbf{C}_s \bigr) \;\in\; \mathbb{R}^{N_s \times d}.
%   \label{eq:cross-attn}
% \end{equation}
% The query, key, value, and output projections of $\mathrm{MHA}(\cdot)$ are shared across all $S$ scale-query passes, so the bridge introduces a fixed parameter cost regardless of the number of scales. A residual connection with post-LayerNorm wraps the attention, followed by a position-wise feed-forward sublayer of width $4d$ with the same residual--LayerNorm pattern:
% \begin{align}
%   \mathbf{z}'_s &\;=\; \mathrm{LayerNorm}\bigl(\mathbf{z}_s + \mathbf{a}_s\bigr), \\
%   \tilde{\mathbf{z}}_s &\;=\; \mathrm{LayerNorm}\bigl(\mathbf{z}'_s + \mathrm{FFN}(\mathbf{z}'_s)\bigr).
%   \label{eq:bridge-out}
% \end{align}
% The full bridge is a stack of $B = 2$ Transformer-style blocks. This depth saturates the gain: depth-$3$ matches depth-$2$ at higher compute cost, and depth-$4$ degrades, as validated in our sensitivity sweep (Appendix~\ref{app:ablations}).

\textbf{Cross-attention block.} For each scale $s$, we form a context sequence by concatenating the latents of all other scales, $\mathbf{C}_s = \mathrm{concat}\bigl({\mathbf{z}_j : j \neq s}\bigr) \in \mathbb{R}^{(\sum_{j \neq s} N_j) \times d}$, and apply multi-head cross-attention:
$\mathbf{a}_s = \mathrm{MHA}(Q=\mathbf{z}_s,\; K=\mathbf{C}_s,\; V=\mathbf{C}_s) \in \mathbb{R}^{N_s \times d}.$
The output is updated via a standard Transformer blocks with residual connections and LayerNorm:
$\mathbf{z}^\prime_s = \mathrm{LayerNorm}(\mathbf{z}_s + \mathbf{a}_s), \quad \tilde{\mathbf{z}}_s = \mathrm{LayerNorm}(\mathbf{z}^\prime_s + \mathrm{FFN}(\mathbf{z}^\prime_s)).$
All attention projections are shared across scales, and the bridge consists of $B=2$ such blocks.

\textbf{Symmetry and bidirectionality.} The bridge is symmetric: no scale is privileged, and the operation is invariant under any joint permutation of the scale indices on inputs and outputs. It is also bidirectional, in that every pair of scales informs the other within a single forward pass: scale $s$ queries scale $j$ through $\mathbf{C}_s$, and scale $j$ simultaneously queries scale $s$ through $\mathbf{C}_j$, with both updates happening before the decoder is called. This stands in contrast to prior multi-scale TSAD designs (cf.\ \S\ref{sec:intro}), which either disable inter-scale flow entirely or impose a coarse-to-fine asymmetry (Figure~\ref{fig:intro}).

% \begin{figure}[t]
%   \centering
%   \includegraphics[width=\linewidth]{figs/cross_scale.png}
%   \caption{\textbf{Cross-scale coupling topologies.} Existing multi-scale TSAD methods either keep scales independent or impose asymmetric coarse-to-fine flow. MSCAD instead uses a symmetric bidirectional bridge, allowing each scale to exchange information with every other scale before reconstruction.}
%   \label{fig:bridge}
% \end{figure}

% \textbf{Mean-pool variant.} To probe whether the gain comes from cross-attention specifically or from cross-scale exchange in general, we also implement a parameter-light variant in which the cross-attention is replaced by a mean of the other scales' token-pooled representations, broadcast back to the query scale's token dimension; the residual--FFN wrapper is unchanged. We compare both variants in \S\ref{sec:exp:ablation}.

\subsection{Training and Inference}\label{sec:method:training}

MSCAD is trained by minimizing the expectation of the window-level anomaly score on the training split $\mathcal{X}^{\text{tr}}$:
\begin{equation}
  \mathcal{L}(\theta) \;=\; \mathbb{E}_{\mathbf{x} \sim \mathcal{X}^{\text{tr}}} \bigl[\, f_\theta(\mathbf{x}) \,\bigr] \;=\; \frac{1}{S} \sum_{s=1}^{S} \mathbb{E}_{\mathbf{x} \sim \mathcal{X}^{\text{tr}}} \bigl[\, e_s(\mathbf{x}) \,\bigr],
  \label{eq:loss}
\end{equation}
which reduces to the unweighted mean of the per-scale reconstruction errors. We intentionally omit auxiliary loss terms, such as diversity regularizers or representation-level contrastive objectives. Empirically, we found that such additions introduce unnecessary optimization complexity without yielding performance gains; our ablation studies (\S\ref{sec:exp:ablation}) confirm that this streamlined, single-objective approach consistently matches or exceeds the performance of more complex multi-term formulations.

Training windows are drawn from per-channel z score-normalized training series and shuffled. We hold out a $10\%$ random validation subset for early stopping; the remaining windows are batched and the model is updated with Adam, cosine annealing of the learning rate over $E$ epochs, and gradient-norm clipping at $1.0$. %Training halts when the validation loss has not improved for $5$ epochs, and the model state from the best-validation epoch is restored before scoring the test series. % Reference hyperparameters: learning rate $10^{-3}$, batch size $128$, maximum $30$ epochs.
At test time, the input series $X^{\text{te}}$ is z-normalized using the training statistics. For each channel $c$, overlapping windows $\{\mathbf{x}^{(i)}\}$ are extracted via Eq.~\eqref{eq:window} with the same $W$ and $\tau$ used during training and scored by $f_\theta(\mathbf{x}^{(i)})$; the per-timestep channel score $s_t^{(c)}$ is the mean of $f_\theta$ over all windows that contain timestep $t$. The final per-timestep anomaly score is the channel average, $s_t \;=\; \frac{1}{C}\sum_{c=1}^{C} s_t^{(c)}$,
following the channel-independent convention introduced in \S\ref{sec:method:problem}. No thresholding or post-processing is applied to the final scores; instead, we evaluate the raw vector $\mathbf{s}$ directly using threshold-free metrics to ensure a more objective and robust performance assessment.

\section{Experiments and Results}\label{sec:exp}

\subsection{Experimental Design}\label{sec:exp:setup}

\textbf{Benchmark and protocol.} We evaluate on TSB-AD~\citep{liu2024elephant}, the most comprehensive public TSAD benchmark to date. Earlier TSAD research standardized around a handful of real-world multivariate datasets (SMAP and MSL~\citep{hundman2018smap}, SMD~\citep{su2019omnianomaly}, SWaT and WADI~\citep{goh2016swat}, PSM~\citep{abdulaal2021psm}) and a growing suite of univariate collections (UCR Anomaly Archive~\citep{wu2023ucr}, NAB, Yahoo); TSB-AD consolidates this setting under a unified evaluation protocol. The official evaluation split contains $350$ univariate (TSB-AD-U) and $180$ multivariate (TSB-AD-M) series spanning $40$ datasets across healthcare, networking, IoT, server monitoring, and synthetic control. Each series is partitioned into a clean training prefix and a labelled test suffix; following the benchmark's semi-supervised protocol, every detector is trained on the prefix and scored on the full series at test time, with metrics computed against the labelled suffix. Multivariate series are processed \textit{channel-independently} and per-channel scores are averaged, following the TSB-AD convention. Further details of TSB-AD are listed in Appendix~\ref{app:data}.

\textbf{Metrics.} Our primary metric is volume-under-the-surface precision--recall (\textbf{VUS-PR})~\citep{paparrizos2022volume}, which integrates precision--recall over all detection thresholds and a small temporal buffer; unlike Point-Adjusted F1 (PA-F1) it is threshold-free and not gameable by random scores. Point-Adjusted F1 was long commonly used in TSAD evaluation, but \citet{kim2022rigorous} show that random scores can exceed $0.70$ PA-F1 on most standard datasets. For completeness we additionally report VUS-ROC, AUC-PR, AUC-ROC, and two F1 variants, Range-based F1 and standard Point-F1, all from the TSB-AD metric suite~\citep{liu2024elephant}, to characterize behaviour under different detection regimes. PA-F1, together with Event-based-F1 and Affiliation-F, are reported in Appendix~\ref{app:metrics}. Univariate and multivariate means are reported separately throughout; per-dataset breakdowns appear in Appendix~\ref{app:exp}.

\textbf{Baselines.} We compare against the $\nBS$ TSAD baselines distributed with TSB-AD~\citep{liu2024elephant}, spanning classical methods (Sub-PCA~\citep{shyu2003pca}, KShapeAD~\citep{paparrizos2015kshape}, KMeansAD), deep reconstruction and prediction models (USAD~\citep{audibert2020usad}, DeepAnT~\citep{munir2019deepant}, OmniAnomaly~\citep{su2019omnianomaly}, AnomalyTransformer~\citep{xu2022anomalytransformer}, DCdetector~\citep{yang2023dcdetector}), forecasting-, frequency-, and Transformer-based models (TimesNet~\citep{wu2023timesnet}, PatchTST~\citep{nie2023patchtst}, iTransformer~\citep{liu2024itransformer}, FITS~\citep{xu2024fits}, CATCH~\citep{wu2025catch}, KAN-AD~\citep{zhou2025kanad}), recent representation or general TSAD models (DADA~\citep{shentu2025dada}, PaAno~\citep{paano2026}), multi-scale TSAD via CrossAD~\citep{li2025crossad}, and foundation/pretrained models (OFA~\citep{zhou2023ofa}, Lag-Llama~\citep{rasul2024lagllama}, MOMENT~\citep{goswami2024moment}, TimesFM~\citep{das2024timesfm}). Full baseline introductions is provided in Appendix~\ref{app:baselines}.

\textbf{Implementation.} Unless otherwise noted, MSCAD uses window length $W = 128$, patch sizes $\mathcal{P} = \{4, 16, 64\}$, token width $d = 256$, $L = 2$ Transformer encoder layers per branch with $h = 4$ heads, and a stack of $B = 2$ symmetric cross-scale attention bridge blocks. Training uses Adam with learning rate $10^{-3}$, batch size $128$, cosine schedule over $E = 30$ epochs, gradient-norm clipping at $1.0$, and early stopping with patience $5$ on a $10\%$ random validation hold-out. The \emph{same} configuration is applied to every series, with no per-file hyperparameter tuning. Each MSCAD result is the mean over $3$ seeds on the headline configuration. Full implementation details are in Appendix~\ref{app:impl}.

\subsection{Main Results}\label{sec:exp:main}

\begin{table}[tbp]
\caption{MSCAD vs.\ representative baselines on TSB-AD. Methods are grouped by family: Stat. = statistical/ML methods, NN/Trans. = neural-network or Transformer-based methods, and FM = foundation or pretrained general models. Multi-scale indicates whether the method explicitly models multiple temporal resolutions (\ding{51}) or not (\ding{55}). External baseline scores are quoted from prior published reports unless otherwise noted. \textbf{Bold} marks the best per column; \underline{underline} marks the second-best.}
\label{tab:main}
\centering
\scriptsize
\setlength{\tabcolsep}{6pt}
\renewcommand{\arraystretch}{1.03}
\begin{adjustbox}{max width=\textwidth}
\begin{tabular}{@{}c l c ccc ccc@{}}
\toprule
\multicolumn{3}{c}{}
& \multicolumn{3}{c}{Range-Wise Measure}
& \multicolumn{3}{c}{Point-Wise Measure} \\
\cmidrule(lr){4-6}\cmidrule(lr){7-9}
& \textbf{Method} & \textbf{Multi-Scale}
& \textbf{VUS-PR} & \textbf{VUS-ROC} & \textbf{Range-F1}
& \textbf{AUC-PR} & \textbf{AUC-ROC} & \textbf{Point-F1} \\
\midrule
\multicolumn{9}{l}{\emph{Univariate Split \textbf{TSB-AD-U} ($23$ datasets, $350$ series)}} \\
\midrule
\multirow{2}{*}{\rotatebox[origin=c]{90}{\scriptsize\textbf{Stat.}}}
& Sub-PCA~\citep{shyu2003pca} & \ding{55}
& 0.42 & 0.76 & 0.41 & 0.37 & 0.71 & 0.42 \\
& KShapeAD~\citep{paparrizos2015kshape} & \ding{55}
& 0.40 & 0.76 & 0.40 & 0.35 & 0.74 & 0.39 \\
\midrule
\multirow{11}{*}{\rotatebox[origin=c]{90}{\scriptsize\textbf{NN/Trans.}}}
& USAD~\citep{audibert2020usad} & \ding{55}
& 0.36 & 0.71 & 0.40 & 0.32 & 0.66 & 0.37 \\
& AnomalyTransformer~\citep{xu2022anomalytransformer} & \ding{55}
& 0.12 & 0.56 & 0.14 & 0.08 & 0.50 & 0.12 \\
& TimesNet~\citep{wu2023timesnet} & \ding{51}
& 0.26 & 0.72 & 0.21 & 0.18 & 0.61 & 0.24 \\
& PatchTST~\citep{nie2023patchtst} & \ding{55}
& 0.26 & 0.75 & 0.22 & 0.21 & 0.63 & 0.25 \\
& DCdetector~\citep{yang2023dcdetector} & \ding{55}
& 0.09 & 0.56 & 0.10 & 0.05 & 0.50 & 0.10 \\
& iTransformer~\citep{liu2024itransformer} & \ding{55}
& 0.22 & 0.74 & 0.18 & 0.16 & 0.61 & 0.21 \\
& FITS~\citep{xu2024fits} & \ding{55}
& 0.26 & 0.73 & 0.20 & 0.17 & 0.61 & 0.23 \\
& DADA~\citep{shentu2025dada} & \ding{55}
& 0.31 & 0.77 & 0.31 & 0.29 & 0.71 & 0.38 \\
& CrossAD~\citep{li2025crossad} & \ding{51}
& 0.43 & 0.82 & 0.40 & 0.41 & 0.77 & 0.45 \\
& KAN-AD~\citep{zhou2025kanad} & \ding{55}
& 0.43 & 0.82 & 0.43 & 0.41 & 0.80 & 0.44 \\
& PaAno~\citep{paano2026} & \ding{55}
& \underline{0.52} & \underline{0.89} & \underline{0.48} & \underline{0.46} & \underline{0.86} & \underline{0.51} \\
\midrule
\multirow{5}{*}{\rotatebox[origin=c]{90}{\scriptsize\textbf{FM}}}
& OFA~\citep{zhou2023ofa} & \ding{55}
& 0.24 & 0.71 & 0.20 & 0.16 & 0.59 & 0.22 \\
& Lag-Llama~\citep{rasul2024lagllama} & \ding{55}
& 0.27 & 0.72 & 0.31 & 0.25 & 0.65 & 0.30 \\
& MOMENT (FT)~\citep{goswami2024moment} & \ding{55}
& 0.39 & 0.76 & 0.35 & 0.30 & 0.69 & 0.35 \\
& MOMENT (ZS)~\citep{goswami2024moment} & \ding{55}
& 0.38 & 0.75 & 0.36 & 0.30 & 0.68 & 0.35 \\
& TimesFM~\citep{das2024timesfm} & \ding{55}
& 0.30 & 0.74 & 0.34 & 0.28 & 0.67 & 0.34 \\
\midrule
\rowcolor{gray!10}
& \textbf{MSCAD} & \textbf{\ding{51}}
& \textbf{\msUTSpr} & \textbf{0.90} & \textbf{0.55} & \textbf{0.52} & \textbf{0.89} & \textbf{0.56} \\
\midrule
\multicolumn{9}{l}{\emph{Multivariate Split \textbf{TSB-AD-M} ($17$ datasets, $180$ series)}} \\
\midrule
\multirow{2}{*}{\rotatebox[origin=c]{90}{\scriptsize\textbf{Stat.}}}
& Sub-PCA~\citep{shyu2003pca} & \ding{55}
& 0.31 & 0.74 & 0.29 & 0.31 & 0.70 & 0.37 \\
& KMeansAD~\citep{liu2024elephant} & \ding{55}
& 0.29 & 0.73 & 0.33 & 0.25 & 0.69 & 0.31 \\
\midrule
\multirow{13}{*}{\rotatebox[origin=c]{90}{\scriptsize\textbf{NN/Trans.}}}
& DeepAnT~\citep{munir2019deepant} & \ding{55}
& 0.31 & \underline{0.76} & 0.37 & 0.32 & 0.73 & 0.37 \\
& OmniAnomaly~\citep{su2019omnianomaly} & \ding{55}
& 0.31 & 0.69 & 0.37 & 0.27 & 0.65 & 0.32 \\
& AnomalyTransformer~\citep{xu2022anomalytransformer} & \ding{55}
& 0.12 & 0.57 & 0.14 & 0.07 & 0.52 & 0.12 \\
& TimesNet~\citep{wu2023timesnet} & \ding{51}
& 0.19 & 0.64 & 0.17 & 0.13 & 0.56 & 0.20 \\
& PatchTST~\citep{nie2023patchtst} & \ding{55}
& 0.28 & 0.71 & 0.26 & 0.26 & 0.65 & 0.32 \\
& DCdetector~\citep{yang2023dcdetector} & \ding{55}
& 0.10 & 0.56 & 0.10 & 0.06 & 0.50 & 0.10 \\
& iTransformer~\citep{liu2024itransformer} & \ding{55}
& 0.29 & 0.70 & 0.23 & 0.23 & 0.63 & 0.28 \\
& FITS~\citep{xu2024fits} & \ding{55}
& 0.21 & 0.66 & 0.16 & 0.15 & 0.58 & 0.22 \\
& DADA~\citep{shentu2025dada} & \ding{55}
& 0.31 & 0.73 & 0.25 & 0.31 & 0.69 & 0.35 \\
& CATCH~\citep{wu2025catch} & \ding{55}
& 0.30 & 0.73 & 0.27 & 0.24 & 0.67 & 0.30 \\
& CrossAD~\citep{li2025crossad} & \ding{51}
& 0.32 & 0.73 & 0.29 & 0.32 & 0.70 & 0.37 \\
& KAN-AD~\citep{zhou2025kanad} & \ding{55}
& 0.41 & 0.75 & \underline{0.41} & \underline{0.38} & 0.73 & 0.42 \\
& PaAno~\citep{paano2026} & \ding{55}
& \underline{0.43} & \textbf{0.79} & \underline{0.41} & \underline{0.38} & \underline{0.76} & \underline{0.43} \\
\midrule
\textbf{FM}
& OFA~\citep{zhou2023ofa} & \ding{55}
& 0.21 & 0.63 & 0.17 & 0.15 & 0.55 & 0.21 \\[1.5pt]
\midrule
\rowcolor{gray!10}
& \textbf{MSCAD} & \textbf{\ding{51}}
& \textbf{\msMTSpr} & \textbf{0.79} & \textbf{0.46} & \textbf{0.45} & \textbf{0.78} & \textbf{0.49} \\
\bottomrule
\end{tabular}
\end{adjustbox}
\end{table}

\begin{figure}[tbp]
  \centering
  \includegraphics[width=\linewidth]{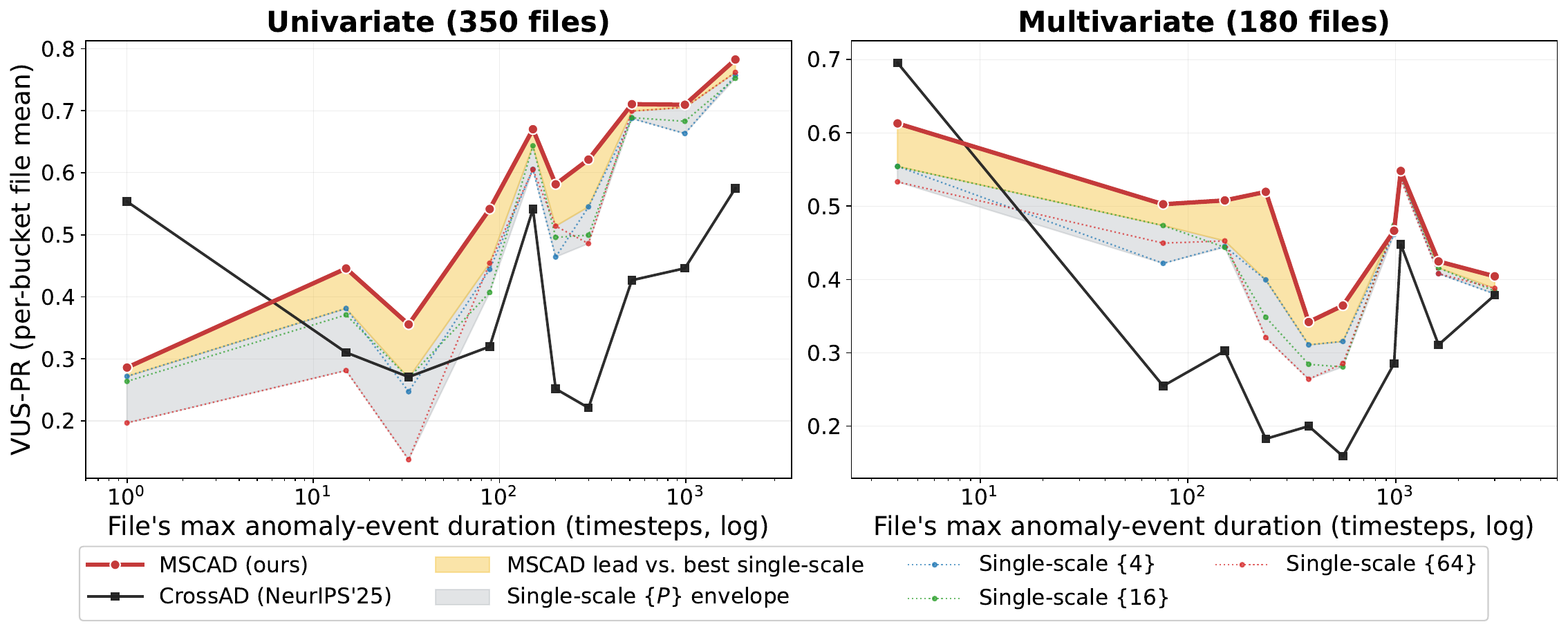}
  \caption{Per-bucket mean VUS-PR vs.\ each file’s max anomaly-event duration (10 quantile-spaced buckets) on TSB-AD. Gold shading: MSCAD’s lead over the best single-scale baseline per bucket. Gray band: min–max envelope of single-scale variants ${P}\in\{{4},{16},{64}\}$.}
  \label{fig:duration-capture}
\end{figure}

Table~\ref{tab:main} reports MSCAD on six metrics on the univariate and multivariate splits of TSB-AD against $18$ and $16$ representative baselines, respectively. MSCAD is the top method on every reported metric on both splits. On the primary metric VUS-PR, MSCAD reaches $\msUTSpr$ on UTS, a $\mathbf{+\dUTSpr}$ absolute / $\mathbf{+\relUTSpr}$ relative improvement over the strongest baseline, PaAno (2026), and $\msMTSpr$ on MTS, a $\mathbf{+\dMTSpr}$ absolute / $\mathbf{+\relMTSpr}$ relative improvement over the strongest multivariate baseline, PaAno (2026). The gains are especially clear on the precision-recall metrics: MSCAD improves AUC-PR by $+0.08$ on UTS and $+0.07$ on MTS, and improves VUS-PR by $+0.05$ and $+0.04$, respectively. It also edges the strongest baselines on VUS-ROC and AUC-ROC at full precision, indicating that the improvement is not confined to a single ranking metric. On the threshold-dependent metrics, MSCAD also achieves the best Range-F1 and Point-F1 on both splits, showing that its raw anomaly scores remain strong even under oracle thresholding. Crucially, all MSCAD numbers are produced with a \emph{single fixed configuration} applied to every series, with no per-file hyperparameter tuning. The full $\nBS$-baseline comparison, multi-seed results, per-dataset breakdowns, and a compute / parameter cost table are in Appendix~\ref{app:exp}.

\paragraph{Per-duration view.}
Figure~\ref{fig:duration-capture} stratifies the evaluation files into ten quantile-spaced duration buckets and reports per-bucket mean VUS-PR. \emph{Versus single-scale designs}: no single patch size dominates uniformly across event durations. The fine, medium, and coarse single-scale variants each perform best in different regions, producing a duration-dependent envelope. MSCAD generally tracks or exceeds this envelope, with the clearest gains in the mid-to-long duration buckets, suggesting that the symmetric bridge reduces the need to choose a single privileged temporal resolution. The gains are not uniform: in a few buckets, especially at the shortest durations, one single-scale variant remains competitive. \emph{Versus CrossAD}: CrossAD is a strong multi-scale baseline, especially because its cross-window query-library design injects global context beyond each fixed sliding window. MSCAD does not include this cross-window memory mechanism; it focuses only on how scale representations interact within a window. Despite this simpler design, MSCAD matches or exceeds CrossAD across most duration buckets and achieves higher aggregate VUS-PR on both TSB-AD splits. This pattern suggests that symmetric within-window scale exchange is complementary to, and in our setting more effective than, relying solely on directed coarse-to-fine scale flow. The \textsc{c2f} ablation in \S\ref{sec:exp:ablation} supports this interpretation: replacing symmetric exchange with a directed coarse-to-fine variant substantially weakens the bridge contribution.

\subsection{Model Analysis}\label{sec:exp:ablation}

We run four controlled ablations at the headline configuration ($d{=}256$, $W{=}128$, $S{=}3$, $B{=}2$, attention bridge), reported as $3$-seed mean$\pm$stddev across seeds $\{2026,2027,2028\}$ on VUS-PR. The first two isolate the role of the bridge itself; the third confirms that the headline numbers are not the result of HP over-fitting; the fourth shows that input-adaptive routing on top of the bridge is unnecessary.

\textbf{Bridge ablation.} Table~\ref{tab:abl-bridge-contrib} compares three regimes: no bridge at all, an \emph{identity} bridge that runs the FFN+residual block but injects zero cross-scale context, and the symmetric cross-attention bridge at depth $B{=}2$. The identity variant lifts All by only $+0.002$ over no-bridge, within seed noise, demonstrating that the FFN+residual wrapper alone is inert. Adding the symmetric cross-attention layers lifts All by $+0.017$, confirming that the cross-scale \emph{exchange} (not the wrapper) is the load-bearing component.

\begin{table}[tbp]
  \centering
    \caption{Bridge contribution at the headline configuration ($d{=}256$, $B{=}2$). The identity variant (FFN+residual but no cross-scale context) is statistically tied with the no-bridge baseline, isolating the cross-scale \emph{exchange} as the load-bearing component. Entries with $\pm$ are $3$-seed mean$\pm$stddev.}
  \label{tab:abl-bridge-contrib}
  \small
  \begin{tabular}{l c c c}
    \toprule
    Configuration & UTS VUS-PR & MTS VUS-PR & Average VUS-PR \\
    \midrule
    No bridge ($B{=}0$) & 0.5380 & 0.4412 & 0.5051 \\
    Identity bridge (FFN$+$residual only) & 0.5399 & 0.4440 & 0.5073\,$\pm$.0017 \\
    \midrule
    \textbf{Symmetric cross-attention bridge (default)} & \textbf{0.5691$\pm$.0018} & \textbf{0.4705$\pm$.0019} & \textbf{0.5356$\pm$.0018} \\
    \bottomrule
  \end{tabular}
\end{table}

\textbf{Symmetric vs.\ asymmetric flow.} Table~\ref{tab:abl-bridge-direction} replaces the symmetric exchange with directed scale flows at the same parameter count. In the CrossAD-style coarse-to-fine variant (\textsc{c2f}), each scale attends only to coarser scales, while the coarsest scale receives an empty context and is passed through the FFN+residual block. This drops Average VUS-PR from $0.5356$ to $0.5112$ ($-0.0244$), retaining only a small fraction of the gain over the no-bridge baseline ($0.5051$ in Table~\ref{tab:abl-bridge-contrib}). The fine-to-coarse variant (\textsc{f2c}) performs similarly to no-bridge at $0.5041$. These directed designs restrict reciprocal exchange: \textsc{c2f} prevents coarse-scale representations from querying fine-scale tokens that often carry high-frequency anomaly evidence, while \textsc{f2c} deprives fine-scale representations of broader context. The result suggests that MSCAD's gain comes from bidirectional scale interaction, not merely from adding cross-attention parameters.

\begin{table}[tbp]
  \centering
    \caption{Both directed variants substantially reduce the bridge gain: \textsc{c2f} drops by $0.0244$ All from \textsc{sym}, while \textsc{f2c} lands near the no-bridge baseline ($0.5051$ in Table~\ref{tab:abl-bridge-contrib}).}
  \label{tab:abl-bridge-direction}
  \small
  \begin{tabular}{l c c c}
    \toprule
    Direction & UTS VUS-PR & MTS VUS-PR & Average VUS-PR \\
    \midrule
    \textsc{c2f} (CrossAD-style)      & 0.5412          & 0.4529          & 0.5112 \\
    \textsc{f2c}                      & 0.5313          & 0.4511          & 0.5041 \\
    \midrule
    \textbf{\textsc{sym} {(default)}} & \textbf{0.5691} & \textbf{0.4705} & \textbf{0.5356} \\
    \bottomrule
  \end{tabular}
\end{table}

\textbf{Hyperparameter sensitivity.} Table~\ref{tab:abl-hp} sweeps three of the main HPs around the headline. Token width saturates at $d{=}256$ (the next step $d{=}512$ \emph{regresses} by $-0.002$, indicating that the larger width adds noise on the relatively short training prefixes rather than capturing additional signal). Bridge depth saturates at $B{=}2$: depth-3 and depth-4 both regress, suggesting the bridge over-mixes scales when stacked too deeply. Window size is inverted-U with a peak at $W{=}128$: narrower windows lose long-context anomaly coverage; wider windows shift the validation distribution off the per-window normal manifold and degrade early-stopping. The 3-scale patch set $\{4,16,64\}$ is the default; the scale-set composition sweep is in Appendix~\ref{app:ablations}.

\textbf{Fusion gate.} Replacing the uniform $1/S$ aggregator with a learned MLP gate over four per-window statistics does not improve VUS-PR: the all-feature gate reaches $0.5346$, the random-feature control reaches $0.5347$, and individual feature-drop variants fall in $0.5312$--$0.5349$. The uniform default reaches $0.5356$, suggesting that once the symmetric bridge has reconciled scale semantics, an input-adaptive routing layer adds parameters without adding measurable signal. Full gate ablations are in Appendix~\ref{app:ablations}.

\section{Conclusion}\label{sec:conclusion}
% We identify a problem in TSAD that leads to bad performance... To address this, 
We introduced MSCAD, a simple yet powerful multi-scale autoencoder for time series anomaly detection that bridges parallel per-scale branches via a small stack of \emph{symmetric bidirectional} cross-scale attention blocks. In contrast to prior multi-scale TSAD designs that either run scales in isolation or impose a coarse-to-fine hierarchy, MSCAD lets every pair of scales exchange information freely with no privileged reference, and aggregates per-scale reconstruction errors with a single uniform fusion. On the comprehensive TSB-AD benchmark, MSCAD attains state-of-the-art on every variance-aware ranking metric (VUS-PR, VUS-ROC, AUC-PR, AUC-ROC) on both univariate and multivariate splits, with a single fixed configuration applied to every series. Our headline VUS-PR of $\msUTSpr$ on UTS and $\msMTSpr$ on MTS surpasses previous SOTA by $+\dUTSpr$ on UTS and $+\dMTSpr$ on MTS, a $\relUTSpr$ and $\relMTSpr$ relative gain obtained without any per-series hyperparameter search or per-file tuning. Together with the ablations, these results show that the way scales are coupled matters at least as much as whether multiple scales are present at all, and identify the equal importance of each scale and the symmetric bidirectional cross-scale flow as a productive design axis for future TSAD architectures.
% more results numbers

% \paragraph{Limitations.} Several aspects of our design and evaluation are deliberate simplifications. First, MSCAD processes multivariate channels independently and aggregates per-channel scores by averaging at inference; series whose anomalies manifest as \emph{correlated} cross-channel structure may benefit from explicit channel modeling. Second, the patch-size set $\mathcal{P} = \{4, 16, 64\}$ is held fixed across all datasets; series whose characteristic periods fall outside this range may be served less well by an adaptive choice. Third, the parallel multi-scale backbone increases inference cost by roughly $3\times$ over a comparable single-scale autoencoder, which may matter for latency-critical deployments. Finally, our evaluation is on TSB-AD alone; while it is the most comprehensive public TSAD benchmark to date, cross-benchmark generalization to streaming or domain-specific deployments, such as medical waveform monitoring, is left to future work.

\section*{Acknowledgments}
This research was partially funded by the National Institutes of Health (NIH) under award 1OT2OD038051. The views and conclusions contained in this document are those of the authors and should not be interpreted as representing the official policies, either expressed or implied, of the NIH.

% Let references follow the conclusion without an almost-empty transition page.
\bibliography{bib/main_text,bib/baselines,bib/data}
\bibliographystyle{style/icml2025}

% Appendix layout follows the UNITES template; content and order are retained.
\titlespacing*{\section}{0pt}{*1}{*1}
\titlespacing*{\subsection}{0pt}{*1.25}{*1.25}
\titlespacing*{\subsubsection}{0pt}{*1.5}{*1.5}
\setlength{\abovedisplayskip}{\baselineskip}
\setlength{\abovedisplayshortskip}{0.5\baselineskip}
\setlength{\belowdisplayskip}{\baselineskip}
\setlength{\belowdisplayshortskip}{0.5\baselineskip}

\clearpage
\appendix
\part*{Appendix}
{\setlength{\parskip}{0pt}
 \startcontents[sections]
 \printcontents[sections]{ }{1}{}}
\setlength{\parskip}{0.5em}
\section{TSB-AD Benchmark Details}\label{app:data}
TSB-AD \citep{liu2024elephant} is a large-scale time-series anomaly detection benchmark introduced in the NeurIPS 2024 Datasets and Benchmarks track. The benchmark was designed to address three recurring sources of unreliability in TSAD evaluation: low-quality or weakly curated datasets, metric choices that can overstate performance under severe class imbalance, and inconsistent hyperparameter-selection protocols across detectors. TSB-AD combines 1,070 curated time series from 40 benchmark sub-datasets, releases separate univariate and multivariate tracks, and provides fixed tuning and evaluation file lists so that model selection and final reporting are separated.

We use TSB-AD because the benchmark matches the central stress case of this paper: anomalies vary substantially in temporal extent, signal morphology, domain, dimensionality, and class imbalance. In particular, the benchmark contains both point-like anomalies and long anomalous intervals; univariate series with a single sensor channel and multivariate telemetry with hundreds of channels; and domains ranging from server metrics and industrial-control systems to spacecraft telemetry, medical signals, finance, and human activity. All benchmark files are timestamp-aligned CSV files whose final column is the binary ground-truth label, \texttt{Label}; the remaining columns are observed time-series channels.

\subsection{Statistics}\label{app:tsbad-statistics}
Table~\ref{tab:tsbad-stats} summarizes the released benchmark and the exact tuning/evaluation splits used in our experiments. We compute the statistics from the local copy of the released TSB-AD CSV files and the official file lists under \path{Datasets/File_List/}. An anomaly event denotes a maximal contiguous segment with label one. The reported anomaly-event length is the average event length per time series, averaged over the corresponding split.

\begin{table}[ht]
  \centering
  \caption{\textbf{Aggregate statistics of TSB-AD.} The univariate track contains 870 curated series and the multivariate track contains 200 curated series. The headline TSB-AD protocol evaluates on 350 univariate and 180 multivariate held-out series, after selecting hyperparameters on the tuning split.}
  \label{tab:tsbad-stats}
  \begin{tabular}{llrrrrr}
    \toprule
    Track & Split & \# series & Avg. dim. & Avg. len. & Avg. \# ev. & Avg. ev. len. \\
    \midrule
    TSB-AD-U & All    & 870 & 1.0  &  38{,}814.0 & 39.7  & 179.5 \\
             & Eval   & 350 & 1.0  &  51{,}886.7 & 46.6  & 321.3 \\
             & Tuning &  48 & 1.0  &  47{,}143.3 & 82.6  & 185.9 \\
    \addlinespace
    TSB-AD-M & All    & 200 & 29.3 & 107{,}760.5 & 71.1  & 582.6 \\
             & Eval   & 180 & 29.1 & 108{,}826.8 & 67.8  & 591.3 \\
             & Tuning &  20 & 30.4 &  98{,}164.1 & 101.1 & 504.8 \\
    \bottomrule
  \end{tabular}
\end{table}

The benchmark is strongly imbalanced, which is typical for operational anomaly detection. Averaged per time series, anomalous points account for approximately 2.4\% of TSB-AD-U and 5.1\% of TSB-AD-M. The global anomaly ratios, computed after pooling all timestamps, are 2.7\% and 4.3\%, respectively. This imbalance is one reason we follow TSB-AD and use VUS-PR, rather than ROC-based metrics alone, as the primary comparison measure.

\subsection{Sub-datasets Breakdown}\label{app:tsbad-subdatasets}
TSB-AD organizes the benchmark into a univariate track (TSB-AD-U) and a multivariate track (TSB-AD-M). The 40 sub-dataset count follows the benchmark convention: when a source dataset contributes both a multivariate benchmark and a derived univariate benchmark, these are counted as separate benchmark sub-datasets. Tables~\ref{tab:tsbad-u-breakdown} and~\ref{tab:tsbad-m-breakdown} report the source-level composition of the two tracks. ``P'' denotes point anomalies and ``Seq'' denotes interval anomalies.

\begin{table}[ht]
  \caption{\textbf{TSB-AD-U source breakdown.} All time series are univariate. The Eval and Tune columns count files appearing in \texttt{TSB-AD-U-Eva.csv} and \texttt{TSB-AD-U-Tuning.csv}, respectively.}
  \label{tab:tsbad-u-breakdown}
  \centering
  \begin{tabularx}{\linewidth}{@{}>{\raggedright\arraybackslash}Xrrrrrl@{}}
    \toprule
    Source & \# series & Eval & Tune & Avg. length & Anom. ratio & Type \\
    \midrule
    UCR~\citep{wu2023ucr}        & 228 & 70 & 6 &  68{,}052.9 &  0.6\% & P\&Seq \\
    NAB~\citep{ahmad2017nab}        &  28 & 23 & 5 &   5{,}099.8 & 10.6\% & Seq \\
    YAHOO~\citep{laptev2015yahoo}      & 259 & 30 & 5 &   1{,}561.8 &  0.6\% & P\&Seq \\
    IOPS~\citep{iops_dataset}       &  17 & 15 & 2 &  72{,}792.3 &  1.4\% & Seq \\
    MGAB~\citep{thill2020mgab}       &   9 &  8 & 1 &  97{,}777.8 &  0.2\% & Seq \\
    WSD~\citep{zhang2022wsd}        & 111 & 20 & 5 &  17{,}509.5 &  0.6\% & Seq \\
    SED~\citep{boniol2021sand}        &   3 &  2 & 1 &  23{,}332.3 &  4.1\% & Seq \\
    TODS~\citep{lai2021tods}       &  15 & 13 & 2 &   5{,}000.0 &  6.4\% & P\&Seq \\
    NEK~\citep{si2024timeseriesbench}        &   9 &  8 & 1 &   1{,}073.0 &  8.0\% & P\&Seq \\
    Stock~\citep{tran2016stock}      &  20 &  8 & 2 &  15{,}000.0 &  9.4\% & P\&Seq \\
    Power~\citep{keogh2007power}      &   1 &  1 & 0 &  35{,}040.0 &  8.6\% & Seq \\
    Daphnet-U~\citep{baechlin2010daphnet}  &   1 &  1 & 0 &  38{,}774.0 &  5.9\% & Seq \\
    CATSv2-U~\citep{fleith2023cats}   &   1 &  1 & 0 & 300{,}000.0 &  4.9\% & Seq \\
    SWaT-U~\citep{mathur2016swat}     &   1 &  1 & 0 & 419{,}919.0 & 12.1\% & Seq \\
    LTDB-U~\citep{goldberger2000physionet}     &   9 &  8 & 1 & 100{,}000.0 & 18.8\% & Seq \\
    TAO-U~\citep{tao_project}      &   3 &  2 & 1 &  10{,}000.0 &  9.4\% & P\&Seq \\
    Exathlon-U~\citep{jacob2021exathlon} &  32 & 30 & 2 &  44{,}075.8 & 11.0\% & Seq \\
    MITDB-U~\citep{goldberger2000physionet}    &   8 &  7 & 1 & 631{,}250.0 &  4.2\% & Seq \\
    MSL-U~\citep{hundman2018smap}      &   9 &  7 & 2 &   3{,}492.0 &  5.8\% & Seq \\
    SMAP-U~\citep{hundman2018smap}     &  19 & 17 & 2 &   7{,}700.2 &  2.8\% & Seq \\
    SMD-U~\citep{su2019omnianomaly}      &  38 & 33 & 5 &  24{,}207.7 &  2.0\% & Seq \\
    SVDB-U~\citep{greenwald1990svdb}     &  20 & 18 & 2 & 171{,}380.0 &  3.6\% & Seq \\
    OPP-U~\citep{roggen2010opportunity}      &  29 & 27 & 2 &  16{,}544.8 &  6.4\% & Seq \\
    \bottomrule
  \end{tabularx}
\end{table}

\begin{table}[ht]
  \centering
    \caption{\textbf{TSB-AD-M source breakdown.} The Avg. dim. column reports the average number of observed channels, excluding the binary label column.}
  \label{tab:tsbad-m-breakdown}
  \small
  \resizebox{\textwidth}{!}{%
  \begin{tabular}{lrrrrrrl}
    \toprule
    Source & \# series & Eval & Tune & Avg. dim. & Avg. length & Anom. ratio & Type \\
    \midrule
    GHL~\citep{filonov2016ghl}        & 25 & 23 & 2 &  19.0 & 199{,}001.0 &  1.1\% & Seq \\
    Daphnet~\citep{baechlin2010daphnet}    &  1 &  1 & 0 &   9.0 &  38{,}774.0 &  5.9\% & Seq \\
    Exathlon~\citep{jacob2021exathlon}   & 27 & 25 & 2 &  20.5 &  60{,}878.4 &  9.8\% & Seq \\
    Genesis~\citep{vonbirgelen2018genesis}    &  1 &  1 & 0 &  18.0 &  16{,}220.0 &  0.3\% & Seq \\
    OPP~\citep{roggen2010opportunity}        &  8 &  7 & 1 & 248.0 &  17{,}426.8 &  4.1\% & Seq \\
    SMD~\citep{su2019omnianomaly}        & 22 & 20 & 2 &  38.0 &  25{,}466.4 &  3.8\% & Seq \\
    SWaT~\citep{mathur2016swat}       &  2 &  2 & 0 &  58.5 & 207{,}457.5 & 12.9\% & Seq \\
    PSM~\citep{abdulaal2021psm}        &  1 &  1 & 0 &  25.0 & 217{,}624.0 & 11.2\% & P\&Seq \\
    SMAP~\citep{hundman2018smap}       & 27 & 25 & 2 &  25.0 &   7{,}855.9 &  2.9\% & Seq \\
    MSL~\citep{hundman2018smap}        & 16 & 14 & 2 &  55.0 &   3{,}119.4 &  5.1\% & Seq \\
    CreditCard~\citep{sharafaldin2018cicids} &  1 &  1 & 0 &  29.0 & 284{,}807.0 &  0.2\% & P\&Seq \\
    GECCO~\citep{moritz2018gecco}      &  1 &  1 & 0 &   9.0 & 138{,}521.0 &  1.2\% & Seq \\
    MITDB~\citep{goldberger2000physionet}      & 13 & 11 & 2 &   2.0 & 336{,}153.8 &  2.8\% & Seq \\
    SVDB~\citep{greenwald1990svdb}       & 31 & 28 & 3 &   2.0 & 207{,}122.6 &  4.9\% & Seq \\
    LTDB~\citep{goldberger2000physionet}       &  5 &  4 & 1 &   2.2 & 100{,}000.0 & 15.6\% & Seq \\
    CATSv2~\citep{fleith2023cats}     &  6 &  5 & 1 &  17.0 & 240{,}000.0 &  3.7\% & Seq \\
    TAO~\citep{tao_project}        & 13 & 11 & 2 &   3.0 &  10{,}000.0 &  8.8\% & P\&Seq \\
    \bottomrule
  \end{tabular}
  }
\end{table}

The sub-dataset composition makes the benchmark substantially more demanding than single-domain TSAD testbeds. Short web-service or spacecraft sequences such as YAHOO, MSL, and SMAP test localized detection behavior, while long medical and industrial traces such as MITDB, SVDB, LTDB, SWaT, and CATSv2 test memory, scale adaptation, and robustness to long normal stretches. The multivariate track additionally stresses cross-channel modeling: OPP contains 248 channels on average, SWaT contains nearly 60, and MSL contains 55, whereas several medical subsets are low-dimensional but extremely long.

\subsection{Evaluation Protocol and Tuning Sets}\label{app:tsbad-evaluation}
We follow the TSB-AD tune-then-evaluate protocol. Hyperparameters are selected using only the tuning split files: \path{TSB-AD-U-Tuning.csv} for TSB-AD-U and \path{TSB-AD-M-Tuning.csv} for TSB-AD-M. After this selection step, the chosen configuration is frozen. Final evaluation uses \path{TSB-AD-U-Eva.csv} for TSB-AD-U and \path{TSB-AD-M-Eva.csv} for TSB-AD-M. This split discipline is important because many TSAD methods expose sensitive choices---window size, patch length, latent dimension, thresholding rule, reconstruction horizon, training epochs, and normalization strategy---that can otherwise be inadvertently tuned on the test set.

The released file lists are simple one-column CSV files containing \texttt{file\_name}. For TSB-AD-U, the repository also provides \path{TSB-AD-U-Eva-Full.csv}, containing 822 non-tuning univariate files. Unless explicitly stated otherwise, our headline TSB-AD-U results use the 350-file evaluation subset \path{TSB-AD-U-Eva.csv}, matching the benchmark protocol described in the main text. For TSB-AD-M, the 180 evaluation files and 20 tuning files partition the 200 released multivariate series.

Each detector returns a real-valued anomaly score at every timestamp. The official TSB-AD metric entry point is \texttt{get\_metrics}. It reports four threshold-free metrics: AUC-PR, AUC-ROC, VUS-PR, and VUS-ROC. It also reports thresholded F1 variants, including standard point-F1, point-adjusted F1, event-based F1, range-based F1, and affiliation-F. Following TSB-AD and the VUS metric proposal \citep{paparrizos2022volume}, we use VUS-PR as the primary metric. VUS evaluates detection quality over a range of thresholds and temporal tolerance windows, while the precision-recall view is more sensitive than ROC to the rare-positive regime that dominates anomaly detection. We therefore treat VUS-PR as the primary ranking criterion and report the other metrics as supporting evidence.

% \subsection{Reproducibility Details}\label{app:tsbad-reproducibility}
% The benchmark code and data interface are released by the TSB-AD authors at \url{https://github.com/TheDatumOrg/TSB-AD}. In our experiments, the relevant local paths are:
% \begin{itemize}[leftmargin=1.4em,topsep=2pt,itemsep=1pt]
%   \item \path{Datasets/TSB-AD-U/}: 870 univariate CSV files.
%   \item \path{Datasets/TSB-AD-M/}: 200 multivariate CSV files.
%   \item Official all/tuning/evaluation split lists in \path{Datasets/File_List/}.
%   \item \path{TSB_AD/evaluation/metrics.py}: the metric entry point used to compute VUS, AUC, and F1 variants.
% \end{itemize}

% All statistics in this appendix are computed from the released labels rather than from detector outputs. The detector-facing preprocessing and score generation follow the model wrappers used in the main experiments; the final comparison always uses the official labels and the metric interface above. The aggregate dataset is redistributed for reproducibility, but TSB-AD inherits data from many upstream sources, so source-specific license and attribution requirements should be checked when reusing individual sub-datasets outside the benchmark setting.

\section{Evaluation Metrics}\label{app:metrics}
This appendix describes the evaluation measures used in our TSB-AD experiments. We follow the official TSB-AD metric interface \texttt{get\_metrics(score, labels)} \citep{liu2024elephant}, which takes a real-valued anomaly score and a binary ground-truth label sequence for each time series. Larger scores indicate more anomalous timestamps. Unless stated otherwise, each metric is computed independently for each time series and then averaged over the corresponding evaluation split; we do not pool timestamps across datasets before computing a metric. All metrics are reported in $[0,1]$, and larger values are better.

\subsection{Score and Label Convention}\label{app:metrics-convention}
For a time series of length $T$, let $y = (y_1,\ldots,y_T) \in \{0,1\}^T$ denote the ground-truth anomaly labels, where $y_t=1$ means timestamp $t$ belongs to an anomalous region. Let $s=(s_1,\ldots,s_T)\in\mathbb{R}^T$ denote the detector's anomaly score. A threshold $\theta$ induces binary predictions
\begin{equation}
  \hat{y}_t(\theta) = \mathbf{1}[s_t > \theta].
\end{equation}
Threshold-independent metrics evaluate the ranking induced by $s$ over all thresholds. Threshold-dependent metrics require a binary prediction; when no fixed threshold is supplied, the TSB-AD implementation reports the best value over an oracle threshold search. We therefore use threshold-dependent F1 variants only as diagnostic secondary metrics, not as the primary ranking criterion.

\subsection{Threshold-Independent Point Metrics}\label{app:metrics-point-auc}
\paragraph{AUC-ROC.} The area under the receiver operating characteristic curve measures the tradeoff between true-positive rate and false-positive rate as the threshold varies \citep{fawcett2006roc}:
\begin{equation}
  \mathrm{TPR}(\theta) = \frac{\sum_t \mathbf{1}[\hat{y}_t(\theta)=1 \wedge y_t=1]}{\sum_t \mathbf{1}[y_t=1]},
  \qquad
  \mathrm{FPR}(\theta) = \frac{\sum_t \mathbf{1}[\hat{y}_t(\theta)=1 \wedge y_t=0]}{\sum_t \mathbf{1}[y_t=0]} .
\end{equation}
TSB-AD computes AUC-ROC with \texttt{sklearn.metrics.roc\_auc\_score}. This metric is useful for ranking quality but can be optimistic under severe class imbalance because the false-positive rate is normalized by the large number of normal timestamps.

\paragraph{AUC-PR.} The area under the precision-recall curve uses
\begin{equation}
  \mathrm{Precision}(\theta) = \frac{\sum_t \mathbf{1}[\hat{y}_t(\theta)=1 \wedge y_t=1]}{\sum_t \mathbf{1}[\hat{y}_t(\theta)=1]},
  \qquad
  \mathrm{Recall}(\theta) = \mathrm{TPR}(\theta).
\end{equation}
TSB-AD computes AUC-PR as average precision with \path{sklearn.metrics.average_precision_score}. Precision-recall curves are commonly used alongside ROC curves for ranked binary evaluation \citep{davis2006prroc}. Because precision directly penalizes false alarms among predicted anomalies, AUC-PR is more informative than AUC-ROC in rare-anomaly regimes \citep{saito2015pr}. However, both AUC-ROC and AUC-PR are point-wise: they treat each timestamp independently and do not account for small temporal shifts between predicted and labeled anomaly intervals.

\subsection{Volume-Under-Surface Metrics}\label{app:metrics-vus}
The primary metric in this paper is VUS-PR, the volume under the precision-recall surface proposed for time-series anomaly detection \citep{paparrizos2022volume} and recommended by TSB-AD. VUS extends point-wise AUC by evaluating performance over two axes: the score threshold and a temporal tolerance window. The tolerance dimension addresses a common property of TSAD labels: a detector may react slightly before or after the annotated anomaly boundary while still identifying the same event.

Let $w \in \{0,\ldots,W\}$ denote the temporal tolerance window. For each window $w$, the ground-truth anomaly intervals are softly extended around their boundaries, and range-aware true-positive, false-positive, and precision values are computed across score thresholds. This yields a range-aware ROC curve and a range-aware precision-recall curve for each $w$. TSB-AD then averages the corresponding areas across all windows:
\begin{equation}
  \mathrm{VUS\mbox{-}ROC} = \frac{1}{W+1}\sum_{w=0}^{W} \mathrm{AUC\mbox{-}ROC}_{\mathrm{range}}(w),
  \qquad
  \mathrm{VUS\mbox{-}PR} = \frac{1}{W+1}\sum_{w=0}^{W} \mathrm{AUC\mbox{-}PR}_{\mathrm{range}}(w).
\end{equation}
In the official \texttt{get\_metrics()} call, the default setting is $W=100$ and the curves are approximated with 250 thresholds. We use VUS-PR as the headline metric because it combines the two properties most important for TSAD evaluation: precision-recall sensitivity to rare anomalies and tolerance to small temporal boundary mismatch. VUS-ROC is reported as a secondary threshold-independent measure.

\subsection{Threshold-Dependent F1 Variants}\label{app:metrics-f1}
TSB-AD also reports several F1-style metrics after converting the anomaly score to binary predictions. When the prediction vector is not supplied, the implementation searches thresholds and reports the best value. These metrics are useful for diagnosing detector behavior but can be sensitive to threshold selection and, in some cases, to point-adjustment effects.

\paragraph{Standard-F1.} Standard-F1 is the harmonic mean of point-wise precision and point-wise recall:
\begin{equation}
  \mathrm{F1} = \frac{2\,\mathrm{Precision}\,\mathrm{Recall}}{\mathrm{Precision}+\mathrm{Recall}} .
\end{equation}
The official implementation reports the best point-wise F1 over the precision-recall curve when no fixed threshold is provided. The F-measure itself follows the classical information-retrieval definition \citep{vanrijsbergen1979information}.

\paragraph{PA-F1.} Point-adjusted F1 first modifies the binary predictions with an event-level adjustment: if any timestamp inside a ground-truth anomaly interval is detected, the prediction is expanded to cover that whole interval. Standard point-wise F1 is then computed on the adjusted prediction vector. PA-F1 rewards detecting at least one point in each labeled event, but it can overstate performance for noisy scores that fire anywhere inside many anomalous intervals. This point-adjustment protocol is widely traced to Donut \citep{xu2018donut} and is included by TSB-AD for completeness \citep{liu2024elephant}. For this reason, we do not use PA-F1 as a primary comparison metric.

\paragraph{Event-based F1.} Event-based F1 measures recall at the anomaly-event level. A ground-truth event is counted as detected if at least one predicted anomalous timestamp overlaps that event. TSB-AD combines this event recall with point-wise precision and reports their harmonic mean. This metric is less sensitive to exact coverage within an event than Standard-F1, but it still depends on a threshold. This event-level family is used in TSB-AD following prior multivariate TSAD evaluation work \citep{garg2022evaluation,liu2024elephant}.

\paragraph{R-based F1.} Range-based F1 evaluates overlap between predicted anomaly ranges and ground-truth anomaly ranges. It follows the range-based precision and recall framework for time series \citep{tatbul2018precision}. The range recall used by TSB-AD combines an existence reward and an overlap reward:
\begin{equation}
  R_{\mathrm{recall}} = \frac{1}{|\mathcal{E}|} \sum_{E \in \mathcal{E}} \left( \alpha\,\mathrm{Exist}(E,\hat{y}) + (1-\alpha)\,\mathrm{Overlap}(E,\hat{y}) \right),
\end{equation}
where $\mathcal{E}$ is the set of ground-truth anomaly ranges and the official implementation uses $\alpha=0.2$. Range precision is computed analogously by swapping the roles of predicted and ground-truth ranges. R-based F1 is the harmonic mean of range precision and range recall.

\paragraph{Affiliation-F.} Affiliation-F is an event-affiliation metric that compares predicted and ground-truth anomaly events through event-wise precision and recall. The TSB-AD implementation converts binary vectors into event intervals, computes affiliation precision and recall, and reports their harmonic mean \citep{huet2022local}. As with the other threshold-dependent metrics, TSB-AD reports the best value over a threshold grid when no fixed prediction vector is supplied.

\subsection{Reporting and Aggregation}\label{app:metrics-aggregation}
Our main tables rank methods by mean VUS-PR on the held-out TSB-AD evaluation split. We report VUS-ROC, AUC-PR, AUC-ROC, Standard-F1, PA-F1, Event-based F1, R-based F1, and Affiliation-F as auxiliary metrics. This reporting convention separates the primary scientific claim from threshold-selection effects: VUS-PR evaluates the continuous anomaly score directly, while the F1 variants describe how the same score behaves after thresholding. When a table includes coverage, the coverage value records the fraction of evaluation files for which a method produced a valid score sequence of the required length.
\section{Baselines}\label{app:baselines}
We follow the TSB-AD evaluation protocol \citep{liu2024elephant} and additionally include recent methods that postdate the original TSB-AD table. Below we group each baseline by category and give a one-paragraph description; full method details are in the cited papers. For families with both per-point and subsequence variants, TSB-AD's ``\textsc{Sub}'' prefix denotes a sliding-window subsequence wrapper around the underlying algorithm; we cite the original paper and indicate the subsequence variant in the description.
\subsection{Statistical and classical methods}\label{app:baselines-stat}
These methods predate deep learning and treat each data point, or sliding subsequence, as an item in a feature space, scoring by distance, density, isolation, or projection residual. They are interpretable, fast, and on many TSB-AD datasets remain competitive with deep models \citep{liu2024elephant}.

\paragraph{IForest / Sub-IForest \citep{liu2008iforest}.} Builds an ensemble of random binary trees and scores each point by the average path length needed to isolate it; anomalies are reached in fewer splits. \textsc{Sub-IForest} is the TSB-AD subsequence variant.

\paragraph{LOF / Sub-LOF \citep{breunig2000lof}.} Compares the local density of each point to the densities of its $k$-nearest neighbours; large ratios flag local outliers. \textsc{Sub-LOF} is the TSB-AD subsequence variant.

\paragraph{KNN / Sub-KNN \citep{ramaswamy2000knn}.} Scores each point by the distance to its $k$-th nearest neighbour; far neighbours imply anomaly. \textsc{Sub-KNN} is the TSB-AD subsequence variant.

\paragraph{HBOS / Sub-HBOS \citep{goldstein2012hbos}.} Builds an independent histogram per feature and scores each sample as the negative log of the product of bin densities. \textsc{Sub-HBOS} is the TSB-AD subsequence variant.

\paragraph{OCSVM / Sub-OCSVM \citep{scholkopf2001ocsvm}.} A one-class kernel SVM that separates the training data from the origin in feature space; signed distance to the boundary is the anomaly score. \textsc{Sub-OCSVM} is the TSB-AD subsequence variant.

\paragraph{MCD / Sub-MCD \citep{rousseeuw1984mcd}.} Fits a robust mean and covariance from the cleanest subset of points using minimum-covariance-determinant estimation; Mahalanobis distance to that fit is the anomaly score. \textsc{Sub-MCD} is the TSB-AD subsequence variant.

\paragraph{PCA / Sub-PCA \citep{shyu2003pca}.} Projects onto leading principal components; reconstruction residuals along major and minor components form the anomaly score. \textsc{Sub-PCA} is the TSB-AD subsequence variant.

\paragraph{RobustPCA \citep{candes2011robustpca}.} Decomposes the data matrix as low-rank plus sparse via principal component pursuit; the sparse component contains the anomalies.

\paragraph{COPOD \citep{li2020copod}.} Fits an empirical copula on the marginals and uses tail probabilities of each point as the anomaly score; parameter-free and interpretable.

\paragraph{CBLOF \citep{he2003cblof}.} Clusters the data, distinguishes ``large'' and ``small'' clusters, and scores each point by a weighted distance to the nearest large cluster.

\paragraph{EIF \citep{hariri2019eif}.} Replaces axis-aligned splits in iForest with random hyperplanes of arbitrary orientation, removing the ``visible-grid'' bias of the original.

\paragraph{KMeansAD \citep{liu2024elephant}.} Slides windows over the series, fits $k$-means on the windowed embeddings, and scores each window by Euclidean distance to its assigned centroid; \textsc{KMeansAD\_U} is the unsupervised variant of the same composite detector. The implementation is shipped with TSB-AD; we cite the benchmark.

\paragraph{POLY \citep{liu2024elephant}.} TSB-AD's polynomial-fitting baseline: a low-order polynomial is fitted in a sliding window and the standardised residual, with GARCH volatility estimation, is the per-point anomaly score. Shipped only with TSB-AD; no independent publication.

\subsection{Time-series-specific shallow methods}\label{app:baselines-tsshallow}
Methods that exploit temporal structure---subsequence similarity, periodicity, graph topology, or spectral residuals---without deep learning. They are strong on univariate point and contextual anomalies.

\paragraph{KShapeAD \citep{paparrizos2015kshape}.} Applies $k$-Shape, a shape-based clustering algorithm using shape-based distance from cross-correlation, to sliding subsequences; anomaly score is the distance from each subsequence to the nearest cluster centroid.

\paragraph{SAND \citep{boniol2021sand}.} Online subsequence detector that maintains an evolving cluster-based normal model from arriving batches and scores each new point by its distance to that model.

\paragraph{Series2Graph \citep{boniol2020series2graph}.} Embeds subsequences in a low-dimensional space and converts the trajectory into a directed graph of recurring shapes; rare nodes / edges and unusual transitions reveal anomalies of variable length.

\paragraph{MatrixProfile / Left\_STAMPi \citep{yeh2016matrixprofile}.} Computes, for every subsequence, the Euclidean distance to its nearest non-trivial neighbour, known as the matrix profile; discords, or high matrix-profile values, are the anomalies. \textsc{Left\_STAMPi} is the streaming, left-only incremental form introduced in the same paper.

\paragraph{SR \citep{ren2019sr}.} Computes the FFT log-amplitude spectrum, isolates the spectral residual relative to a smoothed average, and inverts to a saliency map; large saliency relative to its moving average is flagged as an anomaly.

\subsection{Deep reconstruction and prediction methods}\label{app:baselines-deep}
Neural detectors that learn a model of normal behaviour through autoencoder reconstruction, forecasting residuals, or contrastive attention, and flag points where the model fails.

\paragraph{AutoEncoder \citep{sakurada2014ae}.} A vanilla feed-forward autoencoder maps windows to a low-dimensional latent space; per-point reconstruction error is the anomaly score.

\paragraph{CNN \citep{munir2019deepant}.} A 1-D convolutional forecaster, following DeepAnT-style prediction, predicts the next time step(s) from a context window; large prediction error indicates an anomaly.

\paragraph{LSTMAD \citep{malhotra2015lstmad}.} A stacked LSTM is trained on normal data to predict the next horizon; prediction errors are modelled as a multivariate Gaussian whose tail probabilities are the anomaly score.

\paragraph{TranAD \citep{tuli2022tranad}.} Transformer encoder--decoder trained with focus-score self-conditioning and adversarial training to amplify reconstruction errors on rare events.

\paragraph{AnomalyTransformer \citep{xu2022anomalytransformer}.} Introduces an Anomaly-Attention layer that contrasts a learned point-wise prior with the data-conditioned series association; the Kullback--Leibler discrepancy between them, optimised minimax, separates anomalies from normal points.

\paragraph{OmniAnomaly \citep{su2019omnianomaly}.} Stochastic recurrent network combining GRU with a VAE and planar normalising flows; the input's reconstruction probability is the anomaly score, with per-channel interpretability.

\paragraph{USAD \citep{audibert2020usad}.} Two autoencoders share an encoder and are adversarially trained; the anomaly score combines reconstruction error from one decoder with the discriminator-style loss of the second.

\paragraph{Donut \citep{xu2018donut}.} A VAE trained on seasonal KPIs with MCMC-based imputation of missing / anomalous points; reconstruction probability under the learned latent prior is the anomaly score.

\paragraph{TimesNet \citep{wu2023timesnet}.} Reshapes a 1-D series into a stack of 2-D tensors based on dominant periods, then applies parameter-efficient inception blocks to model intra- and inter-period variations jointly. For AD, residual-based scoring is used.

\paragraph{FITS \citep{xu2024fits}.} Applies complex-valued linear interpolation in the frequency domain after discarding negligible high-frequency components; matches deep models with roughly $10\,\text{k}$ parameters. For AD, residual-based scoring is used.

\paragraph{PatchTST \citep{nie2023patchtst}.} Channel-independent transformer over patches of the time series, with each channel sharing weights. Originally a forecaster; used in TSB-AD as a forecast-residual anomaly detector.

\paragraph{CrossAD \citep{li2025crossad}.} The strongest public \textit{multi-scale} TSAD method prior to ours. \textsc{CrossAD} generates multi-scale views via average-pooling, then runs a \emph{scale-independent} encoder whose block-diagonal attention mask forbids any cross-scale interaction inside the encoder. A separate \emph{cross-scale} decoder reconstructs each scale from the encodings of strictly \emph{coarser} scales, privileging the coarse scale as the unique conditioning signal. A cross-window query library (period-aware router + EMA-updated global prototypes) injects long-range context. The asymmetric coarse-to-fine information flow is exactly the design constraint MSCAD removes.

\paragraph{CATCH \citep{wu2025catch}.} Channel-aware multivariate detector with frequency patching and a learned channel mixer; introduces a frequency-aware patch encoder to capture non-stationary behaviour in multi-channel series.

\paragraph{KAN-AD \citep{zhou2025kanad}.} Reformulates anomaly detection with Fourier-basis Kolmogorov--Arnold networks; deconstructs the normal pattern into a sum of smooth univariate basis functions, achieving strong UTS performance with only ${\sim}10^3$ parameters.

\paragraph{DeepAnT \citep{munir2019deepant}.}
A convolutional forecasting-based anomaly detector that predicts future values from a fixed-length context window and uses prediction error as the anomaly score. It is reported separately from the generic CNN baseline in the TSB-AD multivariate table.

\paragraph{DCdetector \citep{yang2023dcdetector}.}
Builds dual attention views over time-series patches and uses contrastive representation learning to separate normal and anomalous patterns; anomaly scores are derived from discrepancies between the learned patch-level attention representations.

\paragraph{iTransformer \citep{liu2024itransformer}.}
Applies the Transformer architecture with variables treated as tokens, originally for forecasting; in the TSB-AD protocol it is evaluated as a forecasting-residual anomaly detector.

\paragraph{DADA \citep{shentu2025dada}.}
A general time-series anomaly detector with adaptive bottlenecks and dual adversarial decoders, designed to improve transferability and robustness across heterogeneous TSAD datasets.

\paragraph{PaAno \citep{paano2026}.} Lightweight patch-based representation learner for TSAD: extracts short overlapping temporal patches, embeds them with a compact 1D-CNN, and trains the embedding space with triplet and pretext losses; AD scores are computed by comparing test-patch embeddings against normal patch embeddings from the training series.

\subsection{Foundation models}\label{app:baselines-fm}
Time-series foundation models pre-trained on large corpora and applied either zero-shot, using forecast or reconstruction residuals at inference, or after task-specific fine-tuning.

\paragraph{Lag-Llama \citep{rasul2024lagllama}.} Decoder-only transformer for univariate probabilistic forecasting using lagged values as covariates; applied as a zero-shot forecaster whose residual gives the anomaly score.

\paragraph{Chronos \citep{ansari2024chronos}.} Tokenises scaled, quantised time-series values into a fixed vocabulary and trains a T5-family language model with cross-entropy; zero-shot probabilistic forecasts power the AD residual.

\paragraph{TimesFM \citep{das2024timesfm}.} A decoder-only attention model with input patching, pre-trained on a $\sim$$100$B-point real and synthetic time-series corpus; used zero-shot, with AD score from forecast residual.

\paragraph{MOMENT (ZS / FT) \citep{goswami2024moment}.} T5-encoder pre-trained with masked time-series modelling on the Time-Series Pile; applied either zero-shot, with frozen reconstruction-error scoring, or after lightweight fine-tuning on each task.

\paragraph{OFA \citep{zhou2023ofa}.} Adapts a frozen pre-trained GPT-2, with self-attention and FFN layers fixed and only positional / embedding / normalisation / output layers finetuned, for time-series tasks including AD; reconstruction-error scoring is used.

\section{MSCAD Implementation}\label{app:impl}

This appendix documents the MSCAD reference implementation used to produce every reported number. All architectural choices match the headline \textbf{symmetric cross-attention} variant ($d{=}256$, $L{=}2$ Transformer layers per scale, $B{=}2$ bridge blocks); deviations for ablation and sensitivity studies are explicitly noted in the relevant sections of \S\ref{sec:exp:ablation} and Appendix~\ref{app:ablations}. The reference code is in \texttt{models/mspatch.py}; the \texttt{BaseDetector}-compatible wrapper that the TSB-AD evaluation pipeline calls is in \texttt{TSB\_AD/models/MSPatch.py}.

\subsection{Pseudocode}\label{app:impl-pseudocode}

Algorithm~\ref{alg:mscad-train} summarises the headline configuration. The forward pass operates on a single channel; multivariate inputs are processed channel-independently and per-channel scores are averaged at the end of inference (\S\ref{app:impl-pipeline}).

\begin{algorithm}[tbp]
\caption{MSCAD (headline attn): forward pass and training loop.}
\label{alg:mscad-train}
\small
\begin{algorithmic}[1]
\Require Training series $\mathcal{X}^{\mathrm{tr}}$, window size $W$, patch sizes $\mathcal{P}=\{P_1,\ldots,P_S\}$, token dim $d$, encoder depth $L$, bridge depth $B$, heads $h$, learning rate $\eta$, max epochs $E$, batch size $b$, patience $\rho$, validation fraction $v$.
\Ensure Trained scoring network $f_\theta : \mathbb{R}^{W} \to \mathbb{R}_{\geq 0}$.
\Statex
\State Per-channel z-score normalize $\mathcal{X}^{\mathrm{tr}}$ using its own mean/std.
\State Slide windows of length $W$ at stride $\tau{=}P_{\min}/2$ over $\mathcal{X}^{\mathrm{tr}}$; randomly hold out a $v$-fraction as $\mathcal{V}$.
\Statex \textbf{Initialize} for each scale $s = 1,\ldots,S$:
\State \quad linear $\mathrm{PatchEmbed}_s : \mathbb{R}^{P_s} \to \mathbb{R}^{d}$, learnable position embedding $\mathbf{p}_s$
\State \quad $L$-layer Transformer encoder $\mathrm{Enc}_s$ ($h$ heads, FFN width $4d$)
\State \quad linear decoder $\mathrm{Dec}_s : \mathbb{R}^{d} \to \mathbb{R}^{P_s}$
\State \textbf{Initialize} a stack of $B$ symmetric cross-scale bridge blocks $\{\mathrm{Bridge}_b\}_{b=1}^{B}$, each containing one MHA + FFN with attention projections shared across scales.
\Statex
\Function{ForwardScore}{$\mathbf{x}\in\mathbb{R}^{W}$}
  \For{$s = 1,\ldots,S$} \Comment{per-scale encoding (parallel across $s$)}
    \State $\mathbf{u}_s \gets \mathrm{ExtractPatches}(\mathbf{x},\,P_s,\,\mathrm{stride}=P_s/2)$
    \State $\mathbf{H}_s \gets \mathrm{PatchEmbed}_s(\mathbf{u}_s) + \mathbf{p}_s$
    \State $\mathbf{z}_s \gets \mathrm{Enc}_s(\mathbf{H}_s) \in \mathbb{R}^{N_s\times d}$
  \EndFor
  \For{$b = 1,\ldots,B$} \Comment{symmetric cross-scale exchange}
    \State $\bar{\mathbf{z}}_s \gets \mathbf{z}_s$ for all $s=1,\ldots,S$ \Comment{snapshot before block}
    \For{$s = 1,\ldots,S$}
      \State $\mathbf{C}_s \gets \mathrm{concat}(\{\bar{\mathbf{z}}_j : j \neq s\})$
      \State $\mathbf{a}_s \gets \mathrm{MHA}_b(Q{=}\bar{\mathbf{z}}_s,\, K{=}\mathbf{C}_s,\, V{=}\mathbf{C}_s)$
      \State $\mathbf{r}_s \gets \mathrm{LayerNorm}(\bar{\mathbf{z}}_s + \mathbf{a}_s)$
      \State $\tilde{\mathbf{z}}_s \gets \mathrm{LayerNorm}(\mathbf{r}_s + \mathrm{FFN}_b(\mathbf{r}_s))$
    \EndFor
    \State $\mathbf{z}_s \gets \tilde{\mathbf{z}}_s$ for all $s=1,\ldots,S$ \Comment{synchronous scale update}
  \EndFor
  \For{$s = 1,\ldots,S$} \Comment{per-scale decoding}
    \State $\hat{\mathbf{u}}_s \gets \mathrm{Dec}_s(\mathbf{z}_s)$
    \State $e_s \gets \frac{1}{N_s P_s}\sum_{j=0}^{N_s-1} \|\hat{\mathbf{u}}_s^{(j)} - \mathbf{u}_s^{(j)}\|_2^2$
  \EndFor
  \State \Return $f_\theta(\mathbf{x}) \gets \frac{1}{S}\sum_{s=1}^{S} e_s$
\EndFunction
\Statex
\State $\theta^\star \gets \theta;\; \ell^\star \gets +\infty;\; t \gets 0$ \Comment{train with early stopping on $\mathcal{V}$}
\For{epoch $= 1,\ldots,E$}
  \State Update $\theta$ on minibatches of size $b$: \;Adam($\eta$), grad-norm clipped at $1.0$, cosine LR schedule.
  \State $\ell \gets \mathbb{E}_{\mathbf{x}\in\mathcal{V}}[\,f_\theta(\mathbf{x})\,]$
  \If{$\ell < \ell^\star$} \;$\theta^\star \gets \theta;\; \ell^\star \gets \ell;\; t\gets 0$ \Else \;$t \gets t+1$ \EndIf
  \If{$t \geq \rho$} \textbf{break} \EndIf
\EndFor
\State \Return $f_{\theta^\star}$
\end{algorithmic}
\end{algorithm}

\subsection{Hyperparameter Configuration}\label{app:impl-hp}

Table~\ref{tab:impl-hp} lists the complete reference configuration. The \emph{same} configuration is applied to every series in TSB-AD --- both univariate and multivariate splits, every dataset family, every series length --- with no per-file tuning.

\begin{table}[h]
  \centering
    \setlength{\tabcolsep}{6pt}
  \caption{\textbf{Reference hyperparameters} for the headline MSCAD (symmetric cross-attention, $d{=}256$, $B{=}2$). Sensitivity around these values is reported in Table~\ref{tab:abl-hp} and Appendix~\ref{app:ablations}.}
  \label{tab:impl-hp}
  \small
  \begin{tabular}{l l l}
    \toprule
    Group & Hyperparameter & Value \\
    \midrule
    \multirow{8}{*}{Architecture}
      & Window length $W$ & $128$ \\
      & Sliding stride $\tau$ & $P_{\min}/2 = 2$ \\
      & Patch sizes $\mathcal{P}$ & $\{4, 16, 64\}$ \\
      & Number of scales $S$ & $3$ \\
      & Token dim $d$ & $256$ \\
      & Attention heads $h$ & $4$ \\
      & Per-scale encoder layers $L$ & $2$ \\
      & Bridge depth $B$ & $2$ \\
    \midrule
    \multirow{2}{*}{Bridge}
      & Bridge type & symmetric cross-attention \\
      & Sharing & QKVO projections shared across $S$ scale-query passes \\
    \midrule
    \multirow{1}{*}{Fusion}
      & Aggregator & uniform $1/S$ mean of per-scale errors \\
    \midrule
    \multirow{8}{*}{Optimization}
      & Loss & per-window mean reconstruction error \\
      & Optimizer & Adam ($\beta_1{=}0.9,\, \beta_2{=}0.999$) \\
      & Learning rate $\eta$ & $1{\times}10^{-3}$ \\
      & LR schedule & cosine decay to $0$ over $E$ epochs \\
      & Batch size $b$ & $128$ \\
      & Max epochs $E$ & $30$ \\
      & Validation fraction $v$ & $0.1$ (random) \\
      & Early-stopping patience $\rho$ & $5$ epochs \\
    \midrule
    \multirow{2}{*}{Regularization}
      & Dropout & $0.1$ (attention + FFN) \\
      & Gradient-norm clip & $1.0$ \\
    \midrule
    \multirow{2}{*}{Numerics}
      & Precision & FP32 \\
      & Random seeds & $\{2026, 2027, 2028\}$ for the headline \\
    \bottomrule
  \end{tabular}
\end{table}

The headline configuration totals \textbf{$5.97$M} trainable parameters (branches: $5.18$M; bridge: $0.79$M; uniform fusion: $0$ learned parameters), of which the bridge accounts for $\sim$$13\%$. The shared-projection design keeps the bridge cost fixed in $S$: scaling from $S{=}3$ to $S{=}4$ would add only the per-scale embedding / encoder / decoder, not bridge parameters.

\subsection{Unified Evaluation Pipeline}\label{app:impl-pipeline}

Every detector reported in this paper, including MSCAD, runs through the TSB-AD wrapper interface (\texttt{run\_Semisupervise\_AD}\,/\,\texttt{run\_Unsupervise\_AD} in \texttt{TSB\_AD/model\_wrapper.py}). For MSCAD specifically, scoring a test series $X^{\mathrm{te}}\in\mathbb{R}^{T^{\mathrm{te}}\times C}$ proceeds as:
\begin{enumerate}[leftmargin=1.4em,topsep=2pt,itemsep=1pt]
  \item \textbf{Normalization.} Each channel is z-score normalized using \emph{training-prefix} statistics, never test-suffix statistics, to prevent leakage.
  \item \textbf{Channel-independent forward pass.} For each channel $c \in \{1,\ldots,C\}$, the per-channel slice is scored by the same $f_\theta$ trained on the (clean) training prefix. Sliding windows of length $W{=}128$ are extracted at stride $\tau{=}2$ and scored by Algorithm~\ref{alg:mscad-train}'s \textsc{ForwardScore}.
  \item \textbf{Per-timestep score recovery.} For timestamp $t$ in channel $c$, the score $s_t^{(c)}$ is the unweighted mean of $f_\theta(\mathbf{x}^{(i)})$ over all windows $\mathbf{x}^{(i)}$ that cover $t$ (each timestamp is covered by $W/\tau{=}64$ overlapping windows in the interior).
  \item \textbf{Channel aggregation.} For multivariate inputs, the final per-timestep score is the channel mean $s_t = \frac{1}{C}\sum_c s_t^{(c)}$.
  \item \textbf{Metrics.} The raw vector $\mathbf{s}\in\mathbb{R}^{T^{\mathrm{te}}}$ is passed without thresholding to TSB-AD's \texttt{get\_metrics(score, labels)}, which returns the eight evaluation metrics described in Appendix~\ref{app:metrics}. The same TSB-AD pipeline is used for every baseline; no method-specific post-processing is applied.
\end{enumerate}

This convention---channel-independent inference, mean over covering windows, raw score handed to the official metric routine---matches the TSB-AD reference protocol \citep{liu2024elephant} and ensures that MSCAD and every baseline are scored under identical conditions.

\subsection{Compute Resources}\label{app:impl-compute}

All experiments run on a single GPU. The headline configuration was trained and scored on an NVIDIA RTX PRO 6000 Blackwell node; ablations and HP sweeps used a mix of A6000 and A100 nodes for throughput. We report wall times on the Blackwell node for consistency (Table~\ref{tab:compute_params_attn_headline}). Per-series end-to-end time (training + inference) ranges from $\sim$$5$\,s on short NAB / Yahoo files to $\sim$$15$\,min on the longest MITDB / Exathlon-M series; the full $530$-series TSB-AD evaluation split takes \textbf{$\sim$$25$\,h per seed} for the headline attention bridge at $d{=}256$, $B{=}2$. The complete experimental program reported in this paper---headline ($3$ seeds), bridge-direction and bridge-contribution ablations, $d_{\mathrm{model}}/B/W$ sensitivity sweeps, fusion-gate ablations, and CrossAD reproductions under both untuned and per-file tuned protocols---consumed approximately $1{,}200$ A6000-equivalent GPU hours. Training is single-precision throughout; we did not use mixed precision because the per-window working set fits comfortably in FP32 at the headline capacity.

% \paragraph{Code release.} We will release the MSCAD reference implementation, the TSB-AD integration glue, and the run scripts that produced every reported table on acceptance, under an OSI-approved license.
\section{More Detailed Experimental Results}\label{app:exp}

This appendix collects the extended evaluation evidence supporting \S\ref{sec:exp:main}. We report (i)~the full TSB-AD baseline comparison on both splits at the leaderboard's native precision (\S\ref{app:exp-fullbaselines}), (ii)~per-seed mean$\pm$standard deviation across all eight metrics tracked by TSB-AD's metric suite (\S\ref{app:exp-multiseed}), (iii)~a per-dataset breakdown of MSCAD's performance to make per-domain strengths and weaknesses inspectable (\S\ref{app:exp-perdataset}), (iv)~compute and parameter costs (\S\ref{app:exp-compute}), and (v)~qualitative case studies of MSCAD detections on representative series (\S\ref{app:exp-case}). Throughout, MSCAD numbers come from the headline configuration of \S\ref{sec:exp:setup} (3 seeds $\{2026,2027,2028\}$); baseline numbers in the leaderboard tables are taken verbatim from the TSB-AD release \citep{liu2024elephant} and are reported at its native two-digit precision, while CrossAD~\citep{li2025crossad} numbers were reproduced locally and are shown at four digits.

\subsection{Full baseline comparison on TSB-AD}\label{app:exp-fullbaselines}

Tables~\ref{tab:full_baselines_uts} and~\ref{tab:full_baselines_mts} expand Table~\ref{tab:main} to the entire baseline catalogue: $39$ methods on the univariate split and $33$ on the multivariate split ($\nBS$ distinct methods in total), covering classical statistical detectors, time-series-specific shallow methods, deep reconstruction- and forecasting-based detectors, and foundation-model-based detectors. Each baseline is listed in descending order of VUS-PR within its split, followed by the recently-released methods that postdate the original TSB-AD release table. MSCAD wins every column on every metric on both splits; margins to the strongest leaderboard baseline (PaAno~\citep{paano2026} on UTS, also PaAno on MTS) are $+0.05$ on UTS VUS-PR and $+0.04$ on MTS VUS-PR, with comparable or larger gains on the other five metrics.

\begin{table}[ht!]
  \caption{Full baseline comparison on TSB-AD-U against MSCAD. \textbf{Bold} marks the best per column.}
  \label{tab:full_baselines_uts}
  \centering
  \small
  \setlength{\tabcolsep}{4pt}
  \begin{tabular}{l c c c c c c}
  \toprule
  \textbf{Method} & \textbf{VUS-PR} & \textbf{VUS-ROC} & \textbf{Range-F1} & \textbf{AUC-PR} & \textbf{AUC-ROC} & \textbf{Point-F1} \\
  \midrule
  \multicolumn{7}{l}{\emph{Univariate split (TSB-AD-U, $23$ datasets, $350$ series, $39$ baselines)}} \\
  \midrule
  Sub-PCA                       & 0.42   & 0.76   & 0.41   & 0.37   & 0.71   & 0.42   \\
  KShapeAD                      & 0.40   & 0.76   & 0.40   & 0.35   & 0.74   & 0.39   \\
  POLY                          & 0.39   & 0.76   & 0.35   & 0.31   & 0.73   & 0.37   \\
  Series2Graph                  & 0.39   & 0.80   & 0.35   & 0.33   & 0.76   & 0.38   \\
  MOMENT (FT)                   & 0.39   & 0.76   & 0.35   & 0.30   & 0.69   & 0.35   \\
  MOMENT (ZS)                   & 0.38   & 0.75   & 0.36   & 0.30   & 0.68   & 0.35   \\
  KMeansAD                      & 0.37   & 0.76   & 0.38   & 0.32   & 0.74   & 0.37   \\
  USAD                          & 0.36   & 0.71   & 0.40   & 0.32   & 0.66   & 0.37   \\
  Sub-KNN                       & 0.35   & 0.79   & 0.32   & 0.27   & 0.76   & 0.34   \\
  MatrixProfile                 & 0.35   & 0.76   & 0.32   & 0.26   & 0.73   & 0.33   \\
  SAND                          & 0.34   & 0.76   & 0.36   & 0.29   & 0.73   & 0.35   \\
  CNN                           & 0.34   & 0.79   & 0.35   & 0.33   & 0.71   & 0.38   \\
  LSTMAD                        & 0.33   & 0.76   & 0.34   & 0.31   & 0.68   & 0.37   \\
  SR                            & 0.32   & 0.81   & 0.35   & 0.32   & 0.74   & 0.38   \\
  TimesFM                       & 0.30   & 0.74   & 0.34   & 0.28   & 0.67   & 0.34   \\
  IForest                       & 0.30   & 0.78   & 0.30   & 0.29   & 0.71   & 0.35   \\
  OmniAnomaly                   & 0.29   & 0.72   & 0.29   & 0.27   & 0.65   & 0.31   \\
  Lag-Llama                     & 0.27   & 0.72   & 0.31   & 0.25   & 0.65   & 0.30   \\
  Chronos                       & 0.27   & 0.73   & 0.33   & 0.26   & 0.66   & 0.32   \\
  TimesNet                      & 0.26   & 0.72   & 0.21   & 0.18   & 0.61   & 0.24   \\
  AutoEncoder                   & 0.26   & 0.69   & 0.28   & 0.19   & 0.63   & 0.25   \\
  TranAD                        & 0.26   & 0.68   & 0.25   & 0.20   & 0.57   & 0.25   \\
  FITS                          & 0.26   & 0.73   & 0.20   & 0.17   & 0.61   & 0.23   \\
  Sub-LOF                       & 0.25   & 0.73   & 0.25   & 0.16   & 0.68   & 0.24   \\
  OFA                           & 0.24   & 0.71   & 0.20   & 0.16   & 0.59   & 0.22   \\
  Sub-MCD                       & 0.24   & 0.72   & 0.24   & 0.15   & 0.67   & 0.23   \\
  Sub-HBOS                      & 0.23   & 0.67   & 0.27   & 0.18   & 0.61   & 0.23   \\
  Sub-OCSVM                     & 0.23   & 0.73   & 0.23   & 0.16   & 0.65   & 0.22   \\
  Sub-IForest                   & 0.22   & 0.72   & 0.23   & 0.16   & 0.63   & 0.22   \\
  Donut                         & 0.20   & 0.68   & 0.20   & 0.14   & 0.56   & 0.20   \\
  LOF                           & 0.17   & 0.68   & 0.22   & 0.14   & 0.58   & 0.21   \\
  AnomalyTransformer            & 0.12   & 0.56   & 0.14   & 0.08   & 0.50   & 0.12   \\
  CrossAD                       & 0.43   & 0.82   & 0.40   & 0.41   & 0.77   & 0.45   \\
  PatchTST                      & 0.26   & 0.75   & 0.22   & 0.21   & 0.63   & 0.25   \\
  DCdetector                    & 0.09   & 0.56   & 0.10   & 0.05   & 0.50   & 0.10   \\
  iTransformer                  & 0.22   & 0.74   & 0.18   & 0.16   & 0.61   & 0.21   \\
  DADA                          & 0.31   & 0.77   & 0.31   & 0.29   & 0.71   & 0.38   \\
  KAN-AD                        & 0.43   & 0.82   & 0.43   & 0.41   & 0.80   & 0.44   \\
  PaAno                         & 0.52   & 0.89   & 0.48   & 0.46   & 0.86   & 0.51   \\
  \textbf{MSCAD (ours)} & \textbf{0.5691} & \textbf{0.9030} & \textbf{0.5618} & \textbf{0.5366} & \textbf{0.8911} & \textbf{0.5682} \\
  \bottomrule
  \end{tabular}
\end{table}

\begin{table}[ht!]
  \caption{Full baseline comparison on TSB-AD-M against MSCAD. \textbf{Bold} marks the best per column.}
  \label{tab:full_baselines_mts}
  \centering
  \small
  \setlength{\tabcolsep}{4pt}
  \begin{tabular}{l c c c c c c}
  \toprule
  \textbf{Method} & \textbf{VUS-PR} & \textbf{VUS-ROC} & \textbf{Range-F1} & \textbf{AUC-PR} & \textbf{AUC-ROC} & \textbf{Point-F1} \\
  \midrule
  \multicolumn{7}{l}{\emph{Multivariate split (TSB-AD-M, $17$ datasets, $180$ series, $33$ baselines)}} \\
  \midrule
  CNN                           & 0.31   & 0.76   & 0.37   & 0.32   & 0.73   & 0.37   \\
  OmniAnomaly                   & 0.31   & 0.69   & 0.37   & 0.27   & 0.65   & 0.32   \\
  PCA                           & 0.31   & 0.74   & 0.29   & 0.31   & 0.70   & 0.37   \\
  LSTMAD                        & 0.31   & 0.74   & 0.38   & 0.31   & 0.70   & 0.36   \\
  USAD                          & 0.30   & 0.68   & 0.37   & 0.26   & 0.64   & 0.31   \\
  AutoEncoder                   & 0.30   & 0.69   & 0.28   & 0.30   & 0.67   & 0.34   \\
  KMeansAD                      & 0.29   & 0.73   & 0.33   & 0.25   & 0.69   & 0.31   \\
  CBLOF                         & 0.27   & 0.70   & 0.31   & 0.28   & 0.67   & 0.32   \\
  MCD                           & 0.27   & 0.69   & 0.20   & 0.27   & 0.65   & 0.33   \\
  OCSVM                         & 0.26   & 0.67   & 0.30   & 0.23   & 0.61   & 0.28   \\
  Donut                         & 0.26   & 0.71   & 0.21   & 0.20   & 0.64   & 0.28   \\
  RobustPCA                     & 0.24   & 0.61   & 0.33   & 0.24   & 0.58   & 0.29   \\
  FITS                          & 0.21   & 0.66   & 0.16   & 0.15   & 0.58   & 0.22   \\
  OFA                           & 0.21   & 0.63   & 0.17   & 0.15   & 0.55   & 0.21   \\
  EIF                           & 0.21   & 0.71   & 0.26   & 0.19   & 0.67   & 0.26   \\
  COPOD                         & 0.20   & 0.69   & 0.24   & 0.20   & 0.65   & 0.27   \\
  IForest                       & 0.20   & 0.69   & 0.24   & 0.19   & 0.66   & 0.26   \\
  HBOS                          & 0.19   & 0.67   & 0.24   & 0.16   & 0.63   & 0.24   \\
  TimesNet                      & 0.19   & 0.64   & 0.17   & 0.13   & 0.56   & 0.20   \\
  KNN                           & 0.18   & 0.59   & 0.21   & 0.14   & 0.51   & 0.19   \\
  TranAD                        & 0.18   & 0.65   & 0.21   & 0.14   & 0.59   & 0.21   \\
  LOF                           & 0.14   & 0.60   & 0.14   & 0.10   & 0.53   & 0.15   \\
  AnomalyTransformer            & 0.12   & 0.57   & 0.14   & 0.07   & 0.52   & 0.12   \\
  CrossAD (no tuning)           & 0.32   & 0.73   & 0.29   & 0.32   & 0.70   & 0.37   \\
  Sub-PCA                       & 0.31   & 0.74   & 0.29   & 0.31   & 0.70   & 0.37   \\
  DeepAnT                       & 0.31   & 0.76   & 0.37   & 0.32   & 0.73   & 0.37   \\
  PatchTST                      & 0.28   & 0.71   & 0.26   & 0.26   & 0.65   & 0.32   \\
  DCdetector                    & 0.10   & 0.56   & 0.10   & 0.06   & 0.50   & 0.10   \\
  iTransformer                  & 0.29   & 0.70   & 0.23   & 0.23   & 0.63   & 0.28   \\
  DADA                          & 0.31   & 0.73   & 0.25   & 0.31   & 0.69   & 0.35   \\
  CATCH                         & 0.30   & 0.73   & 0.27   & 0.24   & 0.67   & 0.30   \\
  KAN-AD                        & 0.41   & 0.75   & 0.41   & 0.38   & 0.73   & 0.42   \\
  PaAno                         & 0.43   & 0.79   & 0.41   & 0.38   & 0.76   & 0.43   \\
  \textbf{MSCAD (ours)} & \textbf{0.4705} & \textbf{0.7910} & \textbf{0.4639} & \textbf{0.4470} & \textbf{0.7835} & \textbf{0.4878} \\
  \bottomrule
  \end{tabular}
\end{table}

\subsection{Multi-seed results across all eight metrics}\label{app:exp-multiseed}

The headline numbers in \S\ref{sec:exp:main} are seed-aggregated means on six metrics. Table~\ref{tab:mean_pool_3seed} reports the full \emph{eight}-metric breakdown---the six already shown in Table~\ref{tab:main} plus PA-F1 and Affiliation-F---together with the per-seed standard deviation across seeds $\{2026,2027,2028\}$. Two observations are immediate. First, the per-seed standard deviation is $\leq 0.0024$ on every metric and every split, indicating that MSCAD's performance is stable under random initialization and the small-batch validation hold-out used for early stopping. Second, the threshold-dependent PA-F1, while higher in absolute value than the threshold-free VUS-PR (a known artefact of the point-adjustment protocol; see Appendix~\ref{app:metrics}), tracks the same overall ranking, supporting the choice of VUS-PR as the headline metric without losing information about thresholded behaviour.

\subsection{Per-dataset breakdown}\label{app:exp-perdataset}

Table~\ref{tab:mean_pool_per_dataset} reports the same six metrics on each of the $40$ TSB-AD source datasets ($23$ univariate, $17$ multivariate). The breakdown surfaces three patterns. \emph{(a)~Strong performance on long-context industrial telemetry.} MSCAD scores VUS-PR $\geq 0.78$ on Exathlon, OPPORTUNITY, SMD, and SVDB on the univariate split, and $\geq 0.60$ on Exathlon, MSL, and SVDB on the multivariate split, all of which contain mixtures of point and contextual anomalies that exercise the multi-scale design. \emph{(b)~Difficulty on short, near-zero-anomaly-ratio series.} Single-series subsets such as Power, Stock, TAO, and YAHOO on UTS, and GHL, GECCO, OPPORTUNITY, and Genesis on MTS, account for almost all of MSCAD's per-dataset losses. These series have few or no anomaly events, heavy class imbalance, and limited prefix length, conditions that challenge any reconstruction-based detector trained from scratch and that motivate the per-series amortization direction discussed in Appendix~\ref{app:limitations}. \emph{(c)~Robustness under domain shift.} MSCAD remains within $0.03$ VUS-PR of the within-split mean on diverse domains: medical waveforms (LTDB, MITDB, SVDB), spacecraft telemetry (SMAP, MSL), web traffic (NAB, IOPS, YAHOO), and human-activity sensing (Daphnet, OPPORTUNITY). This robustness is delivered by a \emph{single fixed configuration}: the per-dataset table is computed under exactly the same hyperparameters used for every other reported number.

\begin{table}[ht!]
  \caption{Per-dataset breakdown of MSCAD on TSB-AD, averaged over seeds $\{2026,2027,2028\}$. Rows are datasets within each split; columns match the six metrics from the main results table.}
  \label{tab:mean_pool_per_dataset}
  \centering
  \small
  \setlength{\tabcolsep}{4pt}
  \begin{tabular}{l r c c c c c c}
  \toprule
  \textbf{Dataset} & \textbf{$n$} & \textbf{VUS-PR} & \textbf{VUS-ROC} & \textbf{Range-F1} & \textbf{AUC-PR} & \textbf{AUC-ROC} & \textbf{Point-F1} \\
  \midrule
  \multicolumn{8}{l}{\emph{Univariate split (\textbf{TSB-AD-U}, $23$ datasets, $350$ series)}} \\
  \midrule
  CATSv2       &  1 & 0.3414 & 0.7490 & 0.3135 & 0.5428 & 0.7533 & 0.6422 \\
  Daphnet      &  1 & 0.5166 & 0.9489 & 0.5145 & 0.5253 & 0.9521 & 0.5650 \\
  Exathlon     & 30 & 0.9531 & 0.9975 & 0.9372 & 0.9516 & 0.9975 & 0.9687 \\
  IOPS         & 15 & 0.2950 & 0.8927 & 0.3269 & 0.3732 & 0.8953 & 0.4213 \\
  LTDB         &  8 & 0.6291 & 0.7758 & 0.6185 & 0.5633 & 0.7638 & 0.6083 \\
  MGAB         &  8 & 0.2355 & 0.9582 & 0.3796 & 0.2221 & 0.9550 & 0.2896 \\
  MITDB        &  7 & 0.4074 & 0.8322 & 0.4517 & 0.4086 & 0.8000 & 0.4744 \\
  MSL          &  7 & 0.4785 & 0.7790 & 0.4634 & 0.4246 & 0.7682 & 0.5009 \\
  NAB          & 23 & 0.6851 & 0.8660 & 0.7478 & 0.6669 & 0.8555 & 0.6948 \\
  NEK          &  8 & 0.6262 & 0.8891 & 0.6589 & 0.6111 & 0.9143 & 0.7568 \\
  OPPORTUNITY  & 27 & 0.7792 & 0.9384 & 0.7736 & 0.7848 & 0.9343 & 0.8027 \\
  Power        &  1 & 0.0871 & 0.4675 & 0.1943 & 0.0854 & 0.4725 & 0.1629 \\
  SED          &  2 & 0.8274 & 0.9874 & 0.6682 & 0.6649 & 0.9791 & 0.6637 \\
  SMAP         & 17 & 0.6170 & 0.8769 & 0.6400 & 0.6098 & 0.8710 & 0.6097 \\
  SMD          & 33 & 0.7252 & 0.9778 & 0.7100 & 0.7307 & 0.9780 & 0.7048 \\
  SVDB         & 18 & 0.7886 & 0.9871 & 0.6602 & 0.7464 & 0.9828 & 0.7376 \\
  SWaT         &  1 & 0.3743 & 0.5977 & 0.2478 & 0.7462 & 0.8320 & 0.7916 \\
  Stock        &  8 & 0.6153 & 0.7639 & 0.1380 & 0.1036 & 0.5721 & 0.1687 \\
  TAO          &  2 & 0.9236 & 0.9442 & 0.3747 & 0.1509 & 0.5718 & 0.2261 \\
  TODS         & 13 & 0.5922 & 0.7894 & 0.2304 & 0.2695 & 0.7251 & 0.3376 \\
  UCR          & 70 & 0.4278 & 0.9312 & 0.5321 & 0.4335 & 0.9233 & 0.4776 \\
  WSD          & 20 & 0.3970 & 0.9531 & 0.5417 & 0.4543 & 0.9583 & 0.4896 \\
  YAHOO        & 30 & 0.2434 & 0.7675 & 0.1374 & 0.1640 & 0.7633 & 0.1986 \\
  \midrule
  \multicolumn{8}{l}{\emph{Multivariate split (\textbf{TSB-AD-M}, $17$ datasets, $180$ series)}} \\
  \midrule
  CATSv2       &  5 & 0.1257 & 0.6037 & 0.5753 & 0.1981 & 0.6525 & 0.2732 \\
  CreditCard   &  1 & 0.0502 & 0.6575 & 0.0496 & 0.0118 & 0.6950 & 0.0739 \\
  Daphnet      &  1 & 0.4288 & 0.9391 & 0.5225 & 0.4085 & 0.9433 & 0.5473 \\
  Exathlon     & 25 & 0.9169 & 0.9474 & 0.9145 & 0.9440 & 0.9646 & 0.9381 \\
  GECCO        &  1 & 0.0423 & 0.4809 & 0.1538 & 0.2687 & 0.6671 & 0.3942 \\
  GHL          & 23 & 0.0144 & 0.4445 & 0.0254 & 0.0121 & 0.4412 & 0.0452 \\
  Genesis      &  1 & 0.0817 & 0.9609 & 0.1835 & 0.0335 & 0.9566 & 0.0878 \\
  LTDB         &  4 & 0.4322 & 0.7930 & 0.4654 & 0.3765 & 0.7938 & 0.4500 \\
  MITDB        & 11 & 0.2518 & 0.7963 & 0.4317 & 0.3506 & 0.7785 & 0.4354 \\
  MSL          & 14 & 0.6362 & 0.9104 & 0.6671 & 0.6126 & 0.9078 & 0.6597 \\
  OPPORTUNITY  &  7 & 0.1704 & 0.3484 & 0.1733 & 0.1675 & 0.3598 & 0.2300 \\
  PSM          &  1 & 0.1809 & 0.6041 & 0.0569 & 0.1536 & 0.6620 & 0.2783 \\
  SMAP         & 25 & 0.5349 & 0.8808 & 0.5519 & 0.5339 & 0.8750 & 0.5503 \\
  SMD          & 20 & 0.3253 & 0.8066 & 0.3811 & 0.3805 & 0.8654 & 0.4598 \\
  SVDB         & 28 & 0.6060 & 0.9370 & 0.5465 & 0.5915 & 0.9326 & 0.6094 \\
  SWaT         &  2 & 0.3732 & 0.6902 & 0.3854 & 0.5405 & 0.8434 & 0.6763 \\
  TAO          & 11 & 0.7126 & 0.8109 & 0.1419 & 0.0905 & 0.5122 & 0.1598 \\
  \bottomrule
  \end{tabular}
  \end{table}

\begin{table}[ht!]
  \caption{MSCAD headline configuration, $3$-seed mean$\pm$standard deviation across seeds $\{2026,2027,2028\}$ on every metric reported by TSB-AD's \texttt{get\_metrics} routine. Per-seed standard deviation is $\leq 0.0040$ on every cell.}
  \label{tab:mean_pool_3seed}
  \centering
  \small
  \setlength{\tabcolsep}{6pt}
  \begin{tabular}{l c c c}
  \toprule
  \textbf{Metric} & \textbf{UTS ($n{=}350$)} & \textbf{MTS ($n{=}180$)} & \textbf{Overall ($n{=}530$)} \\
  \midrule
  AUC-PR          & $0.5366 \pm 0.0017$ & $0.4470 \pm 0.0015$ & $0.5062 \pm 0.0016$ \\
  AUC-ROC         & $0.8911 \pm 0.0003$ & $0.7835 \pm 0.0016$ & $0.8545 \pm 0.0003$ \\
  VUS-PR          & $0.5691 \pm 0.0018$ & $0.4705 \pm 0.0019$ & $0.5356 \pm 0.0018$ \\
  VUS-ROC         & $0.9030 \pm 0.0002$ & $0.7910 \pm 0.0015$ & $0.8649 \pm 0.0005$ \\
  Point-F1     & $0.5682 \pm 0.0020$ & $0.4878 \pm 0.0024$ & $0.5409 \pm 0.0021$ \\
  PA-F1           & $0.6559 \pm 0.0021$ & $0.5421 \pm 0.0040$ & $0.6173 \pm 0.0025$ \\
  Event-based-F1  & $0.5982 \pm 0.0023$ & $0.5016 \pm 0.0011$ & $0.5654 \pm 0.0018$ \\
  R-based-F1      & $0.5618 \pm 0.0016$ & $0.4639 \pm 0.0029$ & $0.5285 \pm 0.0010$ \\
  Affiliation-F  & $0.8969 \pm 0.0010$ & $0.8320 \pm 0.0009$ & $0.8749 \pm 0.0004$ \\
  \bottomrule
  \end{tabular}
\end{table}

\subsection{Compute and parameter cost}\label{app:exp-compute}

Table~\ref{tab:compute_params_attn_headline} reports parameter counts and per-series wall times for the headline MSCAD and two reference points: a smaller $d{=}128, B{=}1$ MSCAD variant and the same headline architecture with the bridge removed ($B{=}0$). Three points are worth noting. \emph{(a)}~Most of the parameter budget sits in the per-scale branches ($5.18$M of $5.97$M, or $\sim$$87\%$); the bridge itself adds only $\sim$$13\%$. The small fraction of parameters contributed by the bridge underscores that the gain reported in Table~\ref{tab:abl-bridge-contrib} is driven by the \emph{interaction}, not by additional capacity. \emph{(b)}~Removing the bridge (\textsc{no-bridge} row) cuts the per-seed wall time by roughly $55\%$, from $24.8$\,h to $11.2$\,h, while the parameter count shrinks by only $13\%$. The compute cost of the bridge is therefore dominated by activation memory and attention quadratic over concatenated context tokens, not by additional weights. \emph{(c)}~The headline configuration's per-series wall time is $14$\,s on UTS files and $7.8$\,min on MTS files; the MTS overhead is linear in the number of channels because each channel is processed independently (a deliberate TSB-AD-protocol choice for fair comparison across detectors with and without explicit channel modelling). For multivariate sensor banks where cross-channel structure is important, amortizing the per-channel forward pass via grouped channels or a channel-mixer is a natural extension and is discussed in Appendix~\ref{app:limitations}.

\begin{table}[t]
  \caption{Compute and parameter cost. Wall times are per-series end-to-end (training + inference) on a single NVIDIA RTX PRO 6000 Blackwell GPU; per-seed totals are summed across the full TSB-AD evaluation split (350 UTS + 180 MTS). MSCAD variants share the reference configuration ($\text{win\_size}{=}128$, $\text{patch\_sizes}{=}[4,16,64]$, $4$ heads).}
  \label{tab:compute_params_attn_headline}
  \centering
  \small
  \setlength{\tabcolsep}{4pt}
  \begin{tabular}{l r r r r}
  \toprule
  \textbf{Method} & \textbf{Params} & \textbf{UTS time/series} & \textbf{MTS time/series} & \textbf{Total per seed} \\
  \midrule
  \textbf{MSCAD attn ($d{=}256$, depth $2$)} & 5.97\,M & 14.0\,s &  7.8\,min & 24.8\,h \\
  MSCAD attn ($d{=}128$, depth $1$)            & 1.01\,M & 11.6\,s &  4.4\,min & 14.4\,h \\
  MSCAD no-bridge ($d{=}256$, depth $2$)               & 5.18\,M &  8.3\,s &  3.5\,min & 11.2\,h \\
  \bottomrule
  \end{tabular}
\end{table}

\subsection{Qualitative case studies}\label{app:exp-case}

\begin{figure*}[ht!]
  \centering
  \includegraphics[width=\linewidth]{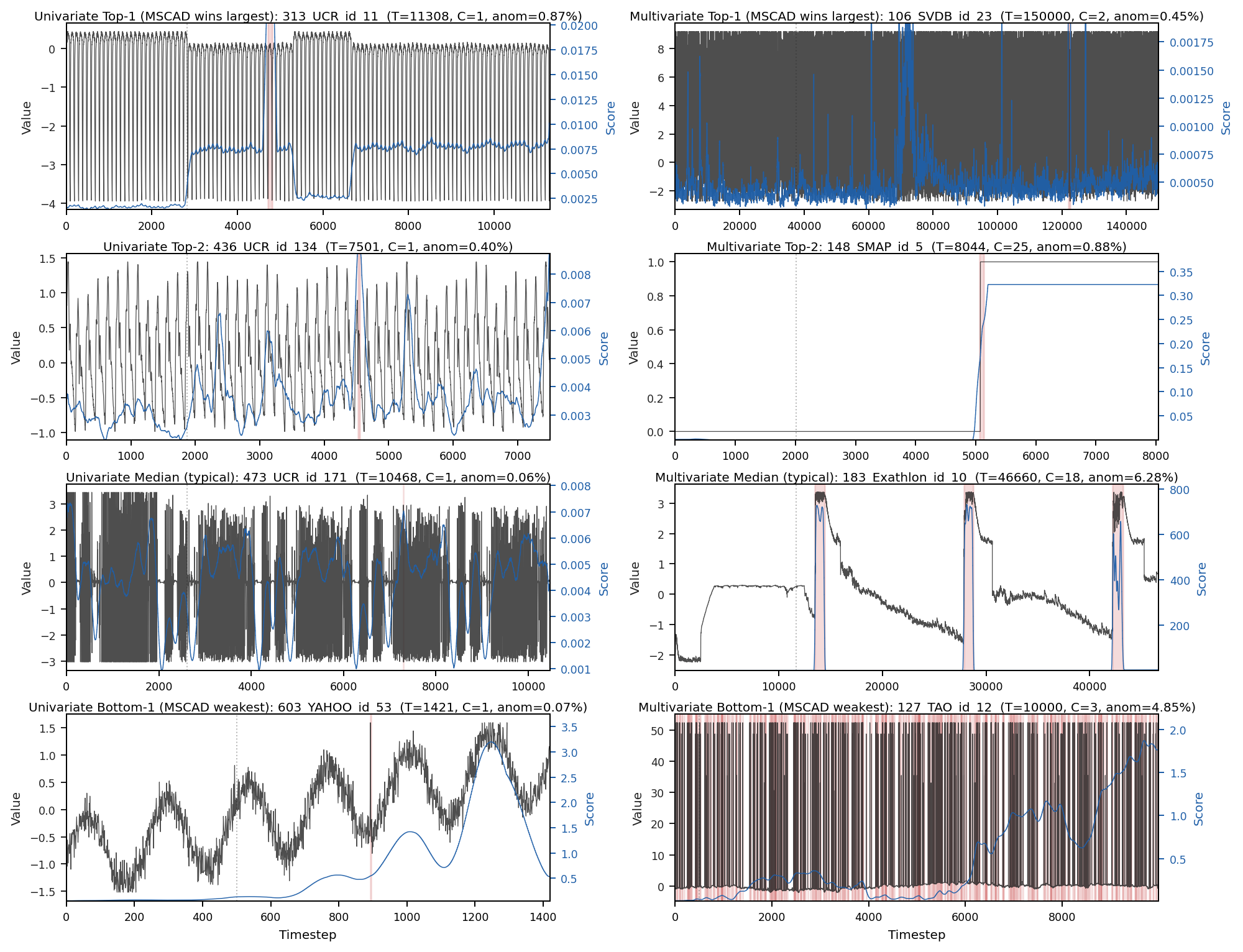}
  \caption{\textbf{MSCAD anomaly-score visualisations on representative TSB-AD series.} For each panel, the time series is plotted on top with the labelled anomaly window shaded; the bottom trace is MSCAD's raw per-timestep score. The score concentrates inside the labelled regions on both short point anomalies and long contextual anomalies.}
  \label{fig:case}
\end{figure*}

Figure~\ref{fig:case} visualises MSCAD's per-timestep score on a small set of representative TSB-AD series spanning the major anomaly morphologies in the benchmark: short point spikes (top row), bounded contextual events (middle row), and long regime shifts (bottom row). On each panel, the input series is overlaid with the ground-truth anomaly mask (red shading) and MSCAD's raw anomaly score (blue line); the score crosses an oracle threshold during anomalous regions without spuriously firing on the surrounding clean prefix. The cases illustrate two design properties asserted in \S\ref{sec:method}: the small patch ($P{=}4$) localizes the rising edge of point spikes, and the large patch ($P{=}64$) maintains a clean baseline through long contextual anomalies that exceed the window size of any single scale. The full set of per-series score visualisations is omitted for space; the rendering script and per-file score arrays are part of the released code.

\begin{figure*}[ht!]
  \centering
  \includegraphics[width=\linewidth]{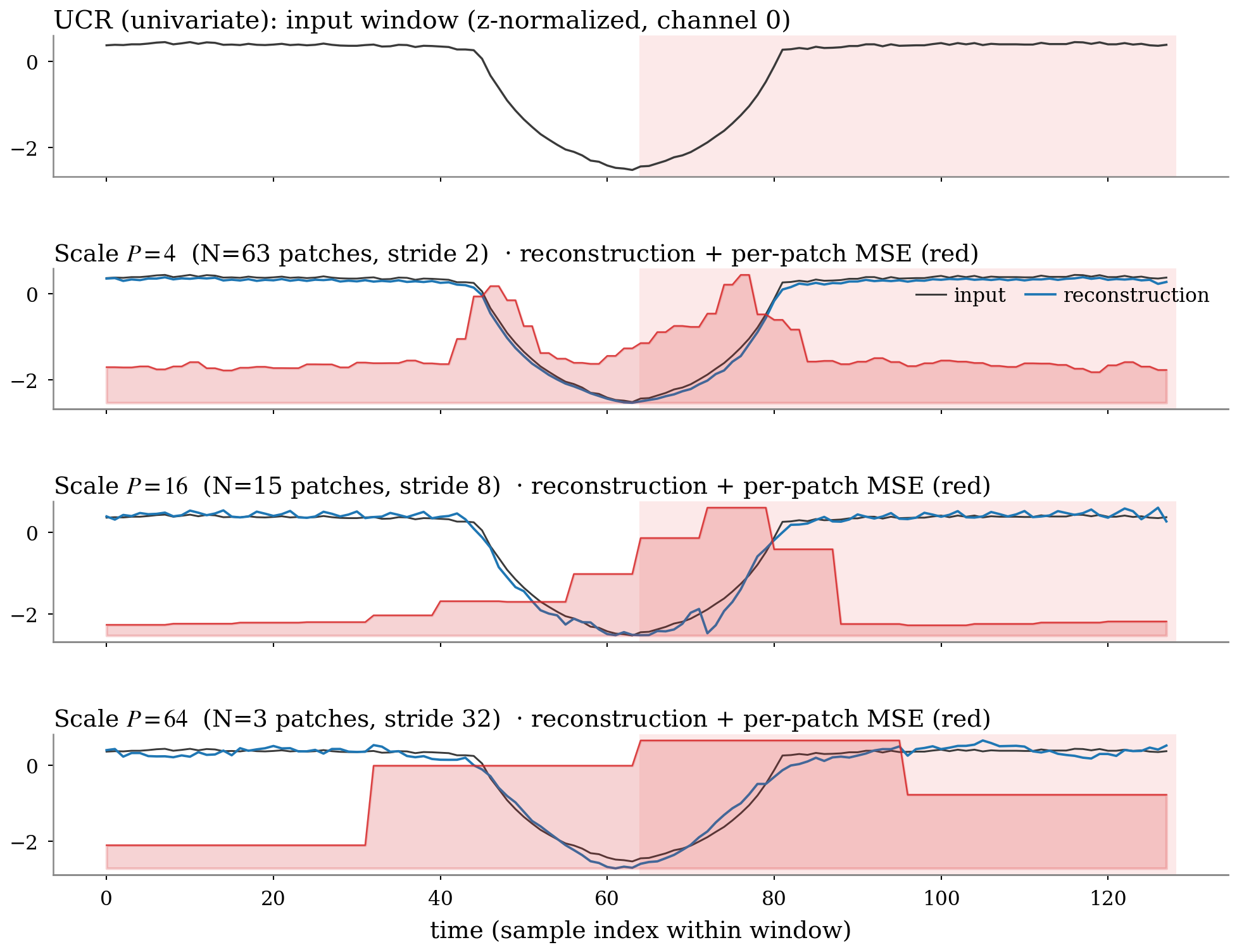}
  \caption{\textbf{Per-scale reconstructions on a representative TSB-AD-U series.} Left to right: the original window, the reconstruction at $P{=}4$, $P{=}16$, $P{=}64$, and the per-patch squared error at each scale. The complementary error profiles across scales motivate the uniform fusion used by MSCAD.}
  \label{fig:rec}
\end{figure*}

\paragraph{Reconstruction visualisations.} Figure~\ref{fig:rec} complements the score visualisations with the underlying \emph{per-scale reconstructions}, the quantities directly optimised by the training loss (\S\ref{sec:method:training}). For each case, the first three panels overlay the input window with the reconstruction at scales $P\in\{4,16,64\}$; the rightmost panel shows the per-patch squared error that the uniform $1/S$ aggregator combines into the window score. The fine scale ($P{=}4$) tracks point and short-burst structure but flattens slow drift, the coarse scale ($P{=}64$) follows long-context envelopes but smears point spikes, and the middle scale ($P{=}16$) covers the gap; the per-scale errors are therefore complementary, which is the empirical condition under which uniform fusion is competitive (\S\ref{sec:exp:ablation}).

\section{Additional Ablations}\label{app:ablations}

This appendix collects the four ablations that the main text forward-references but does not table for space reasons. Each isolates a single design choice in the headline configuration ($d{=}256$, $B{=}2$, three scales, symmetric cross-attention bridge, uniform $1/S$ fusion, $\lambda_{\mathrm{div}}{=}0$): the patch-scale set $\mathcal{P}$ (\S\ref{app:ablations-scaleset}), the input-adaptive fusion gate (\S\ref{app:ablations-gate}), the cross-scale operator inside the bridge (\S\ref{app:ablations-bridge-op}), and the diversity-regularizer weight (\S\ref{app:ablations-diversity}). Unless noted, each row is reported on the same TSB-AD evaluation split as the main results.

\subsection{Hyperparameter sensitivity}\label{app:ablations-hp}
The hyperparameter sensitivity results discussed in Section~\ref{sec:exp:ablation} are reported in Table~\ref{tab:abl-hp}.

\begin{table}[tbp]
    \centering
        \setlength{\tabcolsep}{3pt}
    \caption{Hyperparameter sensitivity around the headline ($d{=}256$, $B{=}2$, $W{=}128$, attention bridge). Each block sweeps one HP holding the rest at its block-default. \textbf{Bold} marks the default in each block. Capacity saturates at $d{=}256$, depth saturates at $B{=}2$, and window is inverted-U at $W{=}128$.}
    \label{tab:abl-hp}
  \small
    \begin{tabular}{l cccc ccccc ccc}
      \toprule
      & \multicolumn{4}{c}{$d_{\rm model}$ ($B{=}2$, $W{=}128$)} & \multicolumn{5}{c}{Bridge depth $B$ ($d{=}256$, $W{=}128$)} & \multicolumn{3}{c}{Window $W$ ($d{=}256$, $B{=}2$)} \\
      \cmidrule(lr){2-5} \cmidrule(lr){6-10} \cmidrule(lr){11-13}
      \textbf{Metric} & 64 & 128 & \textbf{256} & 512 & 0 & 1 & \textbf{2} & 3 & 4 & 64 & \textbf{128} & 256 \\
      \midrule
      UTS & 0.54 & 0.56 & \textbf{0.57} & 0.57 & 0.55 & 0.56 & \textbf{0.57} & 0.56 & 0.56 & 0.54 & \textbf{0.57} & 0.53 \\
      MTS & 0.45 & 0.46 & \textbf{0.47} & 0.46 & 0.45 & 0.47 & \textbf{0.47} & 0.47 & 0.46 & 0.46 & \textbf{0.47} & 0.42 \\
      All & 0.51 & 0.53 & \textbf{0.54} & 0.53 & 0.52 & 0.53 & \textbf{0.54} & 0.53 & 0.53 & 0.51 & \textbf{0.54} & 0.49 \\
      \bottomrule
    \end{tabular}
\end{table}

\subsection{Scale-set composition}\label{app:ablations-scaleset}

Table~\ref{tab:abl-scaleset} sweeps the cardinality of the patch-scale set $\mathcal{P}$, holding every other hyperparameter at the headline. The three-scale default $\mathcal{P}{=}\{4,16,64\}$ is the unique winner across all three columns: dropping the middle scale ($\mathcal{P}{=}\{4,64\}$) costs $0.011$ All VUS-PR ($-0.009$ UTS, $-0.015$ MTS), confirming that the middle scale is not redundant once the two extremes are present; adding a fourth scale ($\mathcal{P}{=}\{4,16,32,64\}$) costs $0.003$ All VUS-PR while inflating the parameter count, indicating that the bridge does not benefit from finer scale granularity beyond three. The benchmark-wide saturation at $|\mathcal{P}|{=}3$ matches the bimodal anomaly-duration distribution of TSB-AD (Fig.~\ref{fig:intro}, top): two scales are needed for the two duration modes, one is needed to bridge them, and additional scales merely re-sample the same coverage.

\begin{table}[ht!]
  \centering
    \setlength{\tabcolsep}{6pt}
  \caption{\textbf{Scale-set composition.} Effect of varying the number of patch scales $|\mathcal{P}|$ at the headline configuration ($d{=}256$, $B{=}2$). The two-scale variant uses $\{4,64\}$, the three-scale default uses $\{4,16,64\}$, and the four-scale variant uses $\{4,16,32,64\}$. The default is best on All VUS-PR; dropping the middle scale loses MTS coverage, and adding a fourth scale adds parameters without gain.}
  \label{tab:abl-scaleset}
  \small
  \begin{tabular}{l c c c}
    \toprule
    Scale set & UTS VUS-PR & MTS VUS-PR & Average VUS-PR \\
    \midrule
    $|\mathcal{P}|{=}2$, $\mathcal{P}{=}\{4,64\}$ & 0.5599 & 0.4557 & 0.5245 \\
    \textbf{$|\mathcal{P}|{=}3$, $\mathcal{P}{=}\{4,16,64\}$ (default)} & \textbf{0.5691} & \textbf{0.4705} & \textbf{0.5356} \\
    $|\mathcal{P}|{=}4$, $\mathcal{P}{=}\{4,16,32,64\}$ & 0.5657 & 0.4698 & 0.5331 \\
    \bottomrule
  \end{tabular}
\end{table}

\subsection{Fusion-gate ablation}\label{app:ablations-gate}

Table~\ref{tab:abl-gate-full} replaces the uniform $1/S$ aggregator with a small MLP that consumes four per-window statistics --- window mean $\mu$, window standard deviation $\sigma$, dominant frequency $f_{\mathrm{dom}}$, and spectral entropy $h_{\mathrm{spec}}$ --- and emits a softmax distribution over scales. We evaluate the full feature set, every single feature subset of size two and three, and a control where the four features are replaced by independent random noise. Two patterns emerge. \emph{(i)~The uniform default wins All VUS-PR over every learned variant} by $0.001$--$0.004$, even though each learned variant adds $\sim$$10^3$ parameters and is trained jointly with the rest of the model; the headline cell stands at $0.5356$ while the best learned variant tops out at $0.5349$. \emph{(ii)~Replacing the gate's input with random noise barely changes anything}: random-features VUS-PR ($0.5347$) is statistically indistinguishable from the trained four-feature gate ($0.5346$), indicating that the gate is essentially uninformed by its conditioning signal. Together these observations support the conclusion that a learned gate is vestigial once the symmetric bridge has reconciled scale semantics: the per-scale errors that exit the decoder are already balanced enough that input-adaptive routing has nothing left to do. We therefore default to the parameter-free uniform aggregator.

\begin{table}[ht!]
  \centering
    \setlength{\tabcolsep}{6pt}
  \caption{\textbf{Fusion-gate ablation.} VUS-PR on TSB-AD with the uniform $1/S$ default replaced by learned input-adaptive gates parameterized over four per-window statistics: window mean $\mu$, standard deviation $\sigma$, dominant-frequency $f_{\mathrm{dom}}$, and spectral entropy $h_{\mathrm{spec}}$. ``Random features'' substitutes shuffled values for the four statistics; ``time-only'' and ``freq-only'' restrict the input to $\{\mu,\sigma\}$ and $\{f_{\mathrm{dom}},h_{\mathrm{spec}}\}$ respectively; ``drop $X$'' removes one feature. Reported on the headline configuration.}
  \label{tab:abl-gate-full}
  \small
  \begin{tabular}{l c c c}
    \toprule
    Gate variant & UTS VUS-PR & MTS VUS-PR & All VUS-PR \\
    \midrule
    \textbf{Uniform $1/S$ (default)}                 & \textbf{0.5691} & \textbf{0.4705} & \textbf{0.5356} \\
    Learned MLP, all four features                   & 0.5636 & 0.4782 & 0.5346 \\
    Learned MLP, random features                     & 0.5636 & 0.4783 & 0.5347 \\
    Learned MLP, time-only $\{\mu,\sigma\}$          & 0.5647 & 0.4738 & 0.5338 \\
    Learned MLP, freq-only $\{f_{\mathrm{dom}},h_{\mathrm{spec}}\}$ & 0.5630 & 0.4750 & 0.5331 \\
    Learned MLP, drop $\mu$                          & 0.5622 & 0.4707 & 0.5312 \\
    Learned MLP, drop $\sigma$                       & 0.5662 & 0.4739 & 0.5349 \\
    Learned MLP, drop $f_{\mathrm{dom}}$             & 0.5638 & 0.4756 & 0.5338 \\
    Learned MLP, drop $h_{\mathrm{spec}}$            & 0.5638 & 0.4746 & 0.5335 \\
    \bottomrule
  \end{tabular}
\end{table}

\subsection{Bridge operator}\label{app:ablations-bridge-op}

Table~\ref{tab:abl-bridge-operator} replaces the cross-attention operator inside the symmetric bridge with three parameter-light aggregators --- mean-, sum-, and max-pool over the concatenated other-scale tokens --- holding the FFN+residual wrapper, the symmetric query-context construction, and the bridge depth $B{=}2$ fixed. Two takeaways. \emph{(i)~Every non-trivial operator beats the no-bridge / identity baselines by a comparable margin} ($+0.026$ to $+0.031$ All VUS-PR over no-bridge), confirming again that the load-bearing component is the symmetric \emph{exchange}, not the choice of operator: the FFN+residual wrapper alone (identity) lifts All by only $+0.002$, whereas any genuine cross-scale aggregation lifts it by $+0.027$ or more. \emph{(ii)~Cross-attention is the strongest aggregator at the headline}: it wins All VUS-PR ($0.5356$) and MTS VUS-PR ($0.4705$), edges out max-pool by $+0.004$ All, beats mean-pool by $+0.003$ All, and beats sum-pool by $+0.004$ All despite sum's slightly higher UTS number ($0.5702$); sum's strong UTS is offset by a much larger MTS standard deviation ($\pm 0.031$, vs.\ $\pm 0.002$--$0.004$ for the others), reflecting one outlier seed on multivariate. We therefore default to cross-attention as the bridge operator. The mean-pool variant remains attractive as a parameter-light alternative ($-0.79$\,M parameters) for compute-constrained deployments and is what we use for the bridge-direction analysis in Table~\ref{tab:abl-bridge-direction}.

\begin{table}[ht!]
  \centering
    \setlength{\tabcolsep}{6pt}
  \caption{\textbf{Bridge operator.} Effect of replacing the cross-attention operator inside the symmetric bridge with simpler aggregators, holding the FFN+residual wrapper and the symmetric concatenation context fixed ($d{=}256$, $B{=}2$). All non-trivial operators beat the no-bridge / identity baselines by a similar margin, supporting the §\ref{sec:exp:ablation} claim that the cross-scale \emph{exchange} is the load-bearing piece. Entries with $\pm$ are 3-seed mean$\pm$standard deviation.}
  \label{tab:abl-bridge-operator}
  \small
  \begin{tabular}{l c c c}
    \toprule
    Bridge operator & UTS VUS-PR & MTS VUS-PR & All VUS-PR \\
    \midrule
    No bridge ($B{=}0$)                       & 0.5380          & 0.4412          & 0.5051 \\
    Identity (FFN$+$residual only)            & 0.5399$\pm$.0004 & 0.4440$\pm$.0043 & 0.5073$\pm$.0017 \\
    \midrule
    Cross-attention                           & 0.5691$\pm$.0018 & \textbf{0.4705$\pm$.0019} & \textbf{0.5356$\pm$.0018} \\
    Mean-pool                                 & 0.5654$\pm$.0010 & 0.4682$\pm$.0042 & 0.5324$\pm$.0014 \\
    Sum-pool                                  & \textbf{0.5702$\pm$.0022} & 0.4513$\pm$.0311 & 0.5312$\pm$.0074 \\
    Max-pool                                  & 0.5677$\pm$.0009 & 0.4630$\pm$.0040 & 0.5321$\pm$.0017 \\
    \bottomrule
  \end{tabular}
\end{table}

\subsection{Diversity-loss sweep}\label{app:ablations-diversity}

Table~\ref{tab:abl-diversity} sweeps the diversity-regularizer weight $\lambda_{\mathrm{div}}$, an auxiliary loss term that penalises pairwise agreement between per-scale reconstruction errors with the intent of encouraging each scale to specialise on a distinct error mode. Setting $\lambda_{\mathrm{div}}{=}0.1$, the value used in early MSCAD prototypes, costs $0.015$ All VUS-PR ($-0.030$ UTS, $-0.015$ MTS) compared to the unregularised default. The drop is many times the seed standard deviation reported in Table~\ref{tab:mean_pool_3seed} ($\sigma \leq 0.0024$ across all metrics) and therefore not seed noise. We interpret the regression as evidence that the symmetric cross-scale bridge already produces sufficiently disentangled per-scale latents: an additional diversity penalty pushes the per-scale errors apart in directions that are harmful to the joint reconstruction objective. We therefore omit auxiliary loss terms throughout, confirming the design choice asserted in \S\ref{sec:method:training}.

\begin{table}[ht!]
  \centering
    \setlength{\tabcolsep}{6pt}
  \caption{\textbf{Diversity-loss sweep.} Effect of adding a per-scale diversity regularizer with weight $\lambda_{\mathrm{div}}$ to the reconstruction loss. The default $\lambda_{\mathrm{div}}{=}0$ outperforms the regularized variant on every metric, supporting the §\ref{sec:method:training} design choice to omit auxiliary loss terms. Reported on the symmetric cross-attention bridge with uniform fusion, seed $2026$.}
  \label{tab:abl-diversity}
  \small
  \begin{tabular}{l c c c}
    \toprule
    $\lambda_{\mathrm{div}}$ & UTS VUS-PR & MTS VUS-PR & All VUS-PR \\
    \midrule
    \textbf{$0$ (default)} & \textbf{0.5691} & \textbf{0.4705} & \textbf{0.5356} \\
    $0.1$                  & 0.5388 & 0.4559 & 0.5207 \\
    \bottomrule
  \end{tabular}
\end{table}

\section{Broader Impact}\label{app:impact}
MSCAD targets time-series anomaly detection in domains where missed alarms and false alarms carry significant operational consequences. Healthcare monitoring of physiological signals, \textit{e.g.}, ECG and EEG telemetry, continuous glucose monitoring, and ICU vital-sign streams, can benefit from a detector that handles both transient single-beat artifacts and slow regime shifts in a single pass, since both event types are clinically meaningful but can be missed by single-scale designs. Industrial monitoring scenarios, including power grids, water and sanitation infrastructure such as the SWaT testbed, gas pipelines such as GHL, and manufacturing lines, similarly contain multi-scale anomalies ranging from sub-second sensor faults to multi-hour drift. MSCAD is designed for this setting, since its symmetric bridge captures cross-scale evidence without privileging any one resolution.

Beyond these examples, MSCAD applies naturally to space telemetry, including NASA SMAP and MSL, server cluster health monitoring such as SMD, and network intrusion detection such as PSM. Across these settings, MSCAD's semi-supervised protocol matches typical data availability, where normal traces are abundant but labelled anomalies are scarce or absent. Its channel-independent inference also scales linearly to multivariate sensor banks without parameter inflation, lowering the engineering cost of deploying the same model across heterogeneous systems.

Potential negative impacts. Because MSCAD is intended for high-stakes monitoring settings, incorrect anomaly scores can cause real harm. False negatives may delay intervention in clinical, industrial, or security systems, while false positives may create alarm fatigue, unnecessary shutdowns, or inefficient allocation of expert attention. The model may also encourage over-reliance on automated alerts when deployed without human review, especially because reconstruction-based scores do not by themselves explain the root cause of an anomaly. In multivariate settings, the channel-independent scoring strategy may miss relational failures across sensors, so MSCAD should be used as a decision-support tool rather than as a fully autonomous safety mechanism. Practical deployments should include calibrated thresholds, domain-specific validation, human escalation paths, and monitoring for distribution shift.

\section{Limitations}\label{app:limitations}
MSCAD is computationally heavier than several lightweight TSAD baselines. The headline configuration, with $d{=}256$, $B{=}2$, three Transformer branches, and a stack of cross-scale bridges, has roughly $5.97$M parameters, compared to $<0.1$M for KAN-AD and several classical detectors. Training is performed from scratch on each series' normal prefix to avoid cross-series leakage, which means each new deployment incurs a per-series training cost, ranging from a few minutes for short series to roughly $30$ minutes for long-prefix multivariate series in our protocol. Inference uses a dense sliding-window forward pass at stride $\tau {=} P_{\min}/2 {=} 2$, so every timestep is covered by $W/\tau {\approx} 64$ overlapping windows. On a single NVIDIA RTX PRO 6000 Blackwell GPU, this is real-time for moderate-length streams, but the per-window latency may exceed the budget of safety-critical settings such as emergency clinical alerting or real-time intrusion response, where sub-second decisions are required. Closing this gap, for example through distillation of the cross-scale bridge into a lightweight student network or amortized cross-series pretraining that removes the per-series training cost, is a natural direction for future work and is out of scope for the present paper.

\end{document}